\documentclass{article}

\usepackage[preprint,nonatbib]{neurips_2025}

\usepackage[numbers,sort]{natbib}
\usepackage[utf8]{inputenc} % allow utf-8 input
\usepackage[T1]{fontenc}    % use 8-bit T1 fonts
\usepackage{xcolor}
\definecolor{neuripsblue}{rgb}{0.21,0.49,0.74}
\usepackage[pagebackref,breaklinks,colorlinks,allcolors=neuripsblue]{hyperref}

\usepackage{graphicx} % for pdf, bitmapped graphics files
\usepackage{amsmath,amssymb,mathtools}
\usepackage{bm}
\usepackage{times} % assumes new font selection scheme installed
\usepackage{algorithm}
\usepackage{algorithmic}
\usepackage{enumitem}
\usepackage{wrapfig}
\usepackage{colortbl,booktabs,multirow}       % tables
\usepackage{xspace}
\usepackage{pifont,wasysym}
\usepackage{extarrows}

\usepackage{listings}
\usepackage{afterpage}
\usepackage{relsize}
\usepackage{sidecap}
\usepackage[nodisplayskipstretch]{setspace}
\usepackage{etoc}
\usepackage{minitoc}
\usepackage{subcaption}

\usepackage{fix-cm} % This package lets LaTeX scale the Computer Modern fonts to non-standard sizes.
\usepackage{anyfontsize} % This package is designed to allow more flexible scaling for fonts, which can sometimes help with symbol fonts that don’t have all sizes available.

\definecolor{myblue}{HTML}{3E3EE9}
\definecolor{myorange}{HTML}{FF9000}
\definecolor{mygreen}{HTML}{39BE28}
\definecolor{myyellow}{HTML}{D1CA00}

\newcommand{\rxmark}{\textcolor{red}{\ding{55}}}
\newcommand{\gcmark}{\textcolor{mygreen}{\ding{51}}}

\newcommand\mypar[1]{\par\vspace{-0.0mm}\noindent\textbf{#1}\;\;}
\newcommand{\ourwork}{\textsc{AgMMU}\xspace}

\title{\textsc{AgMMU}: A Comprehensive Agricultural Multimodal Understanding Benchmark}

\author{%
  Aruna Gauba$^{1,2,5*}$ ~
  Irene Pi$^{1,3,5*}$ ~
  Yunze Man$^{1,4,5}$\textsuperscript{\textdagger} ~
  Ziqi Pang$^{1,4,5}$\textsuperscript{\textdagger} \And 
  Vikram S. Adve$^{1,4,5}$ ~
  Yu-Xiong Wang$^{1,4,5}$ \\ [1.0em]
  $^{1}$University of Illinois Urbana-Champaign ~
  $^{2}$Rice University ~
  $^{3}$Carnegie Mellon University \\ [0.5em]
  $^{4}$AIFARMS ~ $^{5}$Center for Digital Agriculture at UIUC \\ [1.0em]
  $^{*}$\textsuperscript{\textdagger} Equal Contribution ~ \textsuperscript{\textdagger} Project Lead 
}

\begin{document}
\maketitle

\etocdepthtag.toc{mtchapter}
\begin{abstract}
We present \textbf{\textsc{AgMMU}}, a challenging \emph{real‑world} benchmark for evaluating and advancing vision-language models (VLMs) in the knowledge‑intensive domain of agriculture. 
Unlike prior datasets that rely on crowdsourced prompts, \ourwork is distilled from \textit{116{,}231 authentic dialogues} between everyday growers and \textit{USDA-authorized Cooperative Extension experts}. 
Through a three‑stage pipeline: automated knowledge extraction, QA generation, and human verification, we construct (i) \textbf{\textsc{AgMMU}}, an \textit{evaluation set} of 746 multiple‑choice questions (MCQs) and 746 open‑ended questions (OEQs), and (ii) \textbf{\textsc{AgBase}}, a \textit{development corpus} of 57{,}079 multimodal facts covering five high-stakes agricultural topics: insect identification, species identification, disease categorization, symptom description, and management instruction. \ourwork has three key advantages:

\begin{itemize}
    \item \textbf{Authentic \& Expert‑Verified}: All facts, images, and answers originate from real farmer and gardener inquiries answered by credentialed specialists, ensuring high‑fidelity agricultural knowledge.
    \item \textbf{Complete Development Suite}: \ourwork uniquely couples a dual‑format evaluation benchmark (MCQ \emph{and} OEQ) with \textsc{AgBase}, a large‑scale training set, enabling both rigorous assessment and targeted improvement of VLMs.
    \item \textbf{Knowledge‑intensive Challenge}: Our tasks demand the synergy of nuanced visual perception and domain expertise, exposing fundamental limitations of current general‑purpose models and charting a path toward robust, application‑ready agricultural AI.
\end{itemize}

Benchmarking 12 leading VLMs reveals pronounced gaps in fine‑grained perception and factual grounding. Open‑sourced models trail after proprietary ones by a wide margin. Simple fine‑tuning on \textsc{AgBase} boosts open-sourced model performance on challenging OEQs for up to \mbox{11.6\%} on average, narrowing this gap and also motivating future research to propose better strategies in knowledge extraction and distillation from \textsc{AgBase}.
We hope \ourwork stimulates research on domain‑specific knowledge integration and trustworthy decision support in agriculture AI development.

\vspace{1mm}
\textbf{Code}: \url{https://github.com/AgMMU/AgMMU}

\textbf{Data}: \url{https://huggingface.co/datasets/AgMMU/AgMMU_v1}

\end{abstract}   
\newpage
\begin{center}
    \centering
    \captionsetup{type=figure}
    \vspace{-8mm}
    \includegraphics[width=0.98\textwidth]{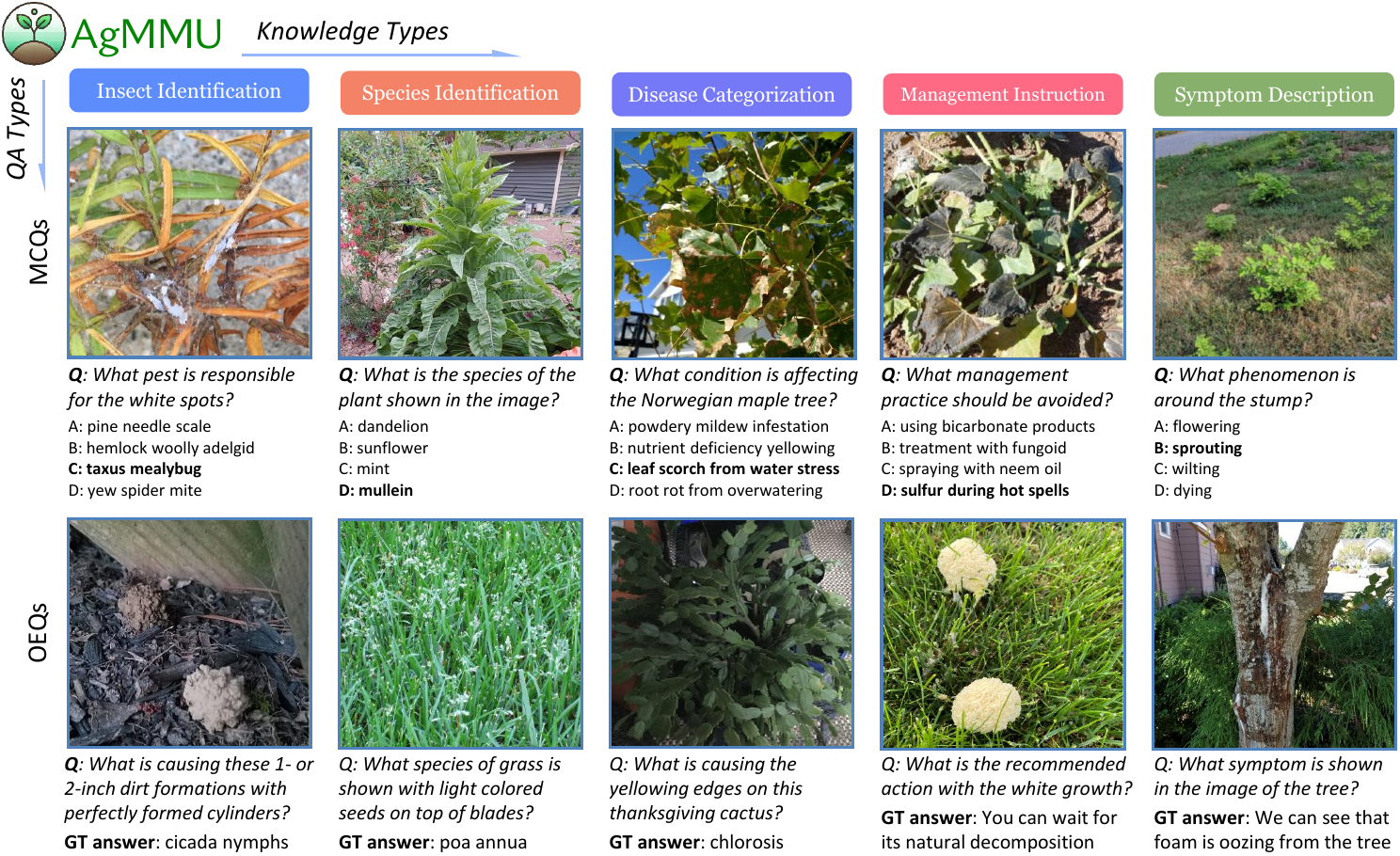}
    \vspace{-1mm}
    \captionsetup{type=figure}
    \captionof{figure}{\textbf{\ourwork} is a multimodal agricultural dataset that challenges vision-language models (VLMs) to observe the details of images and provide factually precise answers. Derived from real-world conversations between users and authorized experts by USDA-funded Cooperative Extension, \ourwork covers five major agricultural knowledge types (demonstrated in five columns of the figure). \ourwork features 746 \textit{multiple-choice questions} (MCQs) like conventional vision-language benchmarks~\cite{yue2024mmmu} and the same number of \textit{open-ended questions} (OEQs) like SimpleQA~\cite{wei2024simpleqa}, all validated by human annotators. We also curate an agricultural knowledge base with 57,079 pieces of information for foundation model fine-tuning, extracted from experts' answers. \ourwork can benefit both knowledge-intensive VLMs and the social good of agriculture.}
    \vspace{-2mm} 
    \label{fig:teaser}
\end{center}
\vspace{-3mm}
\section{Introduction}
\label{sec:introduction}
\vspace{-3mm}
Recent progress in large language models (LLMs) and vision language models (VLMs) has demonstrated remarkable capabilities in general knowledge understanding, as exemplified by the outstanding performance on a variety of multimodal understanding benchmarks~\cite{yue2024mmmu,textvqa,hudson2019gqa,wei2024simpleqa,saikh2022scienceqa,mathew2021docvqa}. However, existing benchmarks often emphasize general-domain tasks and may not fully capture the limitations of current models in more \textit{specialized}, \textit{knowledge-intensive} domains.

Agriculture, a cornerstone application of scientific biological knowledge, presents a particularly challenging domain and has been extensively investigated in the vision and machine learning community since the early years~\cite{gong2024bioscan, maruf2024vlm4bio,stevens2024bioclip,gharaee2024bioscan,inat21,InsectDataSet:2024}. Unlike common tasks in the general domain, agricultural problems often require both precise visual analysis (\textit{e.g.}, identifying specific diseases, pests, or plant conditions) and extensive domain knowledge (\textit{e.g.}, treatment protocols, growing conditions, and management practices). Visual data is exceptionally crucial for agriculture, because many serious problems (\textit{e.g.}, pests, disease, nutrient stress, water stress) can only be recognized visually from nuanced appearances, such as the color and shape of leaves. The stakes are also high in this scenario: accurate and timely agricultural diagnostics can mean the difference between crop survival and failure, directly impacting food security and farmer livelihoods. Despite the critical nature of this domain, \textit{we lack comprehensive benchmarks to evaluate and improve the agricultural understanding abilities of VLMs}, especially the \emph{concerns from real users}.

The challenge of creating such benchmarks is multifaceted. First, biological and agricultural tasks, where computer vision is often applied, are notably labor-intensive for data collection. They require expert-labeled data specific to the target task. Meanwhile, agricultural expertise is scarce and specialized, making it difficult to curate high-quality evaluation data~\cite{cieslak2024generating}. Second, real-world agricultural problems are inherently multimodal requiring both visual and textual understanding, combining the visual observation with background information. Third, no clear protocol defines a representative distribution of realistic agricultural questions. These challenges have left a significant gap in our understanding of how well current AI systems can handle real-world agricultural problems~\cite{FBNNorm:URL,balaguer2024ragvsfinetuningpipelines,Taranis:URL,BayerAITool:URL,DigitalGreen:Farmer.Chat:URL,CropWizard:URL}.

In this work, we introduce the \textbf{\ourwork} (Agricultural Multimodal Understanding Benchmark), the \textit{first} \emph{real-world derived agricultural benchmark} designed to evaluate the capabilities of multimodal foundation models. Our benchmark leverages 116,231 real-world conversations between 2013-2024 hosted by Cooperative Extension~\cite{extensionforum}, which offers one-to-one conversations between \emph{real-world users} capturing images from their own devices and \emph{professional agricultural experts funded by USDA to provide answers}. This real-world data source ensures that our benchmark captures authentic challenges which farmers and gardeners face every day, including complex visual symptoms, varied environmental conditions, and nuanced management decisions.

For data curation, we design an automatic pipeline to extract the agricultural knowledge (as in Fig.~\ref{fig:dataset_curation}) from long-form expert answers and then hire human annotators to verify the quality of the final question-answer pairs. Based on our observation of the real-world questions, the final dataset covers five major types of agricultural questions and knowledge: insect identification, species identification, disease categorization, management instruction, and symptom description, as in Fig.~\ref{fig:teaser}.

Evaluations of a diverse collection of VLMs have revealed that existing foundation models struggle with knowledge-intensive agricultural scenarios, suffering mainly from insufficient knowledge. (refer to Sec.~\ref{sec:analysis} for more details). Moreover, with our preparation of a large multimodal knowledge base, the open-sourced model can be notably improved with plain fine-tuning, even exceeding strong closed-source models in almost all question subdomains. Such observations signify the value of our agricultural benchmark and also encourage exploration of better strategies than simple fine-tuning. 

Our contributions include: 
\begin{enumerate}[leftmargin=*]
    \item \textbf{\textsc{AgMMU}}: A carefully curated evaluation set of 746 multiple-choice questions (MCQs) and the same number of open-ended questions (OEQs, where no candidate choices are provided) extracted from real-world conversations between users and USDA-funded experts and verified by our human annotators. Each extracted QA is designed to test both visual understanding and knowledge application in agricultural contexts.
    \item \textbf{\textsc{AgBase}}: A comprehensive agricultural multimodal knowledge dataset of 57,387 pieces of facts extracted from the long-form experts answers, coming from non-overlapping conversations from the evaluation set. This is suitable for improving VLM performance on agricultural tasks. 
    \item A systematic evaluation of the leading VLMs reveals their limitations in handling knowledge-intensive agricultural queries, along with an error analysis. In the future, \textbf{\ourwork} will provide a complete training and evaluation suite for investigating such problems. 
\end{enumerate}
\vspace{-1mm}
\section{Related Work}
\label{sec:related_work}
\vspace{-1mm}
\mypar{Multimodal Foundation Models and Benchmarks.} 
The evolution of multimodal foundation models has been accompanied by increasingly sophisticated evaluation benchmarks~\cite{li2023llavamedtraininglargelanguageandvision} concentrating on a evolving set of problems significant for AI applications. Early benchmarks like VQAv2~\cite{balanced_vqa_v2} and GQA~\cite{hudson2019gqa} focus on basic visual question-answering capabilities. ScienceQA~\cite{saikh2022scienceqa} specifically targeted scientific reasoning with visual components, while Eyes-Wide-Shut~\cite{eyeswideshut} evaluates the model's ability to avoid hallucination. Recent benchmarks have emphasized real-world applications and complex reasoning. RealWorldQA~\cite{grok} tests the ability of the models to handle practical, everyday visual queries. The MMMU~\cite{yue2024mmmu} benchmark represents a comprehensive effort to evaluate multimodal understanding across multiple domains and task types. On top of these benchmarks, \textsc{AgMMU} presents the first real-world oriented vision-language benchmark targeting agriculture, a domain that faces severe data scarcity and lacks domain experts for large-scale dataset curation.

\mypar{AI in Agriculture.} AI has been extensively applied in the biological and agricultural domains~\cite{TRIPATHI2020183,ZHANG2024109587,PATRICIO201869,awais2024agrogptefficientagriculturalvisionlanguage, zhang2024comprehensive, zhang2024empowering, zhou2024agribench}, with significant datasets and benchmarks driving progress in species identification and disease classification~\cite{10.1145/3371158.3371196,agronomy11112107,9325065,10.1145/3664647.3680599, liu2024multimodal}, answering crop science questions~\cite{zhang2024empowering, silva2023gpt}, and knowledge retrieval~\cite{balaguer2024rag}. In terms of the multimodal perspective of agriculture, the iNaturalist dataset~\cite{inat21} marked a milestone by providing millions of species observations from citizen scientists, enabling the creation of a dataset and a deep neural network model (InsectNet) for the identification of pests~\cite{chiranjeevi:arXiv23,InsectDataSet:2024}. This was followed by more specialized collections like BioScan-1M/5M~\cite{gharaee2024bioscan}, which focused on microscopic biological images, and TreeOfLife-10M~\cite{stevens2024bioclip}, which expands the data in a comprehensive phylogenetic framework. These datasets have facilitated numerous advances in automated species identification, disease detection, and biological image analysis. With the advance of vision-language models (VLMs) in recent years and farmers' need for direct analysis of images, multimodal agricultural benchmarks emerge~\cite{zhang2024empowering, liu2024multimodal} and evaluate the capabilities of VLMs to address agricultural problems. However, a major limitation of these benchmarks is that their images and questions are curated from agricultural experts instead of real-world users with agricultural difficulties, which cannot capture the distribution and complexities of real-world plant growing. From this perspective, our \ourwork fills the gap by building from the conversations between real-world users and experts.

\begin{table*}
\centering
\vspace{-4mm}
\begin{subtable}[t]{0.52\linewidth}
\resizebox{0.99\linewidth}{!}{
\centering
\begin{tabular}{lc@{\hspace{3mm}}c@{\hspace{3mm}}c@{\hspace{3mm}}c@{\hspace{3mm}}c@{\hspace{3mm}}}
\toprule
Datasets & Type & Multimodal & Training & Expert & Factuality
\\

\midrule

% iNat21~\cite{inat21} & CLS   & \rxmark & \gcmark & \gcmark &  -  \\

% TreeOfLife~\cite{stevens2024bioclip}  & CLS  & \gcmark & \gcmark & \gcmark & - \\
SimpleQA~\cite{wei2024simpleqa} & OEQ & \rxmark & \rxmark & \rxmark & \gcmark \\
ScienceQA~\cite{saikh2022scienceqa} & MCQ & \gcmark & \gcmark & \gcmark & \rxmark \\
MMMU~\cite{yue2024mmmu} & MCQ   & \gcmark & \rxmark & \rxmark & \rxmark \\

\midrule
\rowcolor{gray!10}
Our AgMMU  & MCQ+OEQ    & \gcmark & \gcmark & \gcmark & \gcmark \\
\bottomrule
\end{tabular}}\vspace{-1mm}
\caption{Comparison with general benchmarks.}
\end{subtable}%
\begin{subtable}[t]{0.475\linewidth}
\resizebox{0.99\linewidth}{!}{
\centering
\begin{tabular}{lc@{\hspace{3mm}}c@{\hspace{3mm}}c@{\hspace{3mm}}c@{\hspace{3mm}}c@{\hspace{3mm}}}
\toprule
Datasets & Type & Multimodal & Real-world & Factuality
\\
\midrule
iNat21~\cite{inat21} & CLS   & \rxmark & \rxmark &  -  \\
CROP~\cite{zhang2024empowering} & MCQ & \rxmark & \rxmark & \rxmark \\
% TreeOfLife~\cite{stevens2024bioclip}  & CLS  & \gcmark & \gcmark & \gcmark & - \\
% SimpleQA~\cite{wei2024simpleqa} & OEQ & \rxmark & \rxmark & \rxmark & \gcmark \\
% MMMU~\cite{yue2024mmmu} & MCQ   & \gcmark & \rxmark & \rxmark & \rxmark \\
CDDM~\cite{liu2024multimodal} & OEQ & \gcmark & \rxmark & \gcmark \\
\midrule
\rowcolor{gray!10}
Our AgMMU  & MCQ+OEQ    & \gcmark & \gcmark & \gcmark \\
\bottomrule
\end{tabular}}\vspace{-1mm}
\caption{Comparison with agricultural benchmarks.}
\end{subtable}

\vspace{-2mm}
\caption{Objective of AgMMU. (a) AgMMU provides a \emph{multimodal} benchmark in \emph{expert} domains with a \emph{training} set. With open-ended questions (\emph{OEQs}), AgMMU emphasizes the \emph{factual} accuracy of models, where a model has to recall precise facts without relying on options. (b) Besides, AgMMU is unique in leveraging questions from \emph{real-world} users instead of recruited annotators. These properties make AgMMU a unique benchmark for advancing the VLMs for agriculture. }
\label{tab:dataset-comparison}
\vspace{-4mm}  
\end{table*}

\vspace{-3mm}
\section{The \ourwork Benchmark}
\label{sec:method}
\vspace{-3mm}

\subsection{Overview}
\vspace{-2mm}
\mypar{Objective.} We build \ourwork, short for ``Agricultural Multimodal Understanding,'' targeting the major challenges for VLMs to serve users in agricultural domains: \emph{how can VLMs provide precise factual knowledge for real-world questions}? This objective leads to two critical design choices for our data curation. (1) \textbf{Real-world distribution.} The questions, images, and answers are derived from conversations between real users and experts, rather than being curated from the web or textbooks by annotators. (2) \textbf{Factual questions.} To evaluate the factual precision of models, we follow SimpleQA's~\cite{wei2024simpleqa} style by requiring the VLMs to answer open-ended questions (OEQs) in short phrases, directly mentioning the key knowledge helpful for the users. In addition to open-ended questions, we provide multiple-choice questions (MCQs) to align with most multimodal benchmarks like MMMU~\cite{yue2024mmmu}. Our OEQs and MCQs combination provides a comprehensive evaluation. 

\mypar{Data source.} We carefully select the data source to accomplish the above objectives. \ourwork is curated from 116,231 real-world conversations between 2013-2024 hosted by Cooperative Extension~\cite{extensionforum}. Such a data source offers the following critical advantages: (1) \textbf{Real-world users}: the questions and images are posted by real users who have agricultural questions; (2) \textbf{Authorized experts}: USDA funds the cooperative extension to authorize a group of professional agricultural experts from universities to answer these questions. Only certified experts are allowed to provide answers, which ensures the quality of data; (3) \textbf{Eligibility for research}: With the Cooperative Extension funded by USDA and being a non-profitable organization, the data can be open-sourced for research purposes under the CC-BY-SA4.0 License. Pre-processing is executed to remove personal and confidential information from our release dataset.

\mypar{AgMMU Overview.} We design pipelines involving both automatic processing and human verification to turn raw conversations into high-quality QA pairs for agriculture knowledge evaluation. After filtering and processing, our \ourwork comprises of: (1) an evaluation set with 746 MCQ and the same number of OEQ questions covering the major agricultural knowledge types (as in Fig.~\ref{fig:teaser}); (2) a larger scale development set with 57,079 QA pairs formatted as open-ended QA, which is disjoint with the evaluation set and aimed to support the fine-tuning of VLMs for agricultural knowledge. We compare \ourwork with related AI and agricultural benchmarks in Table~\ref{tab:dataset-comparison}, marking its uniqueness.

\vspace{-3mm}
\subsection{Dataset Curation}
\label{sec:data_curation}
\begin{figure*}
    \centering
    \vspace{-4mm}
    \includegraphics[width=1\linewidth]{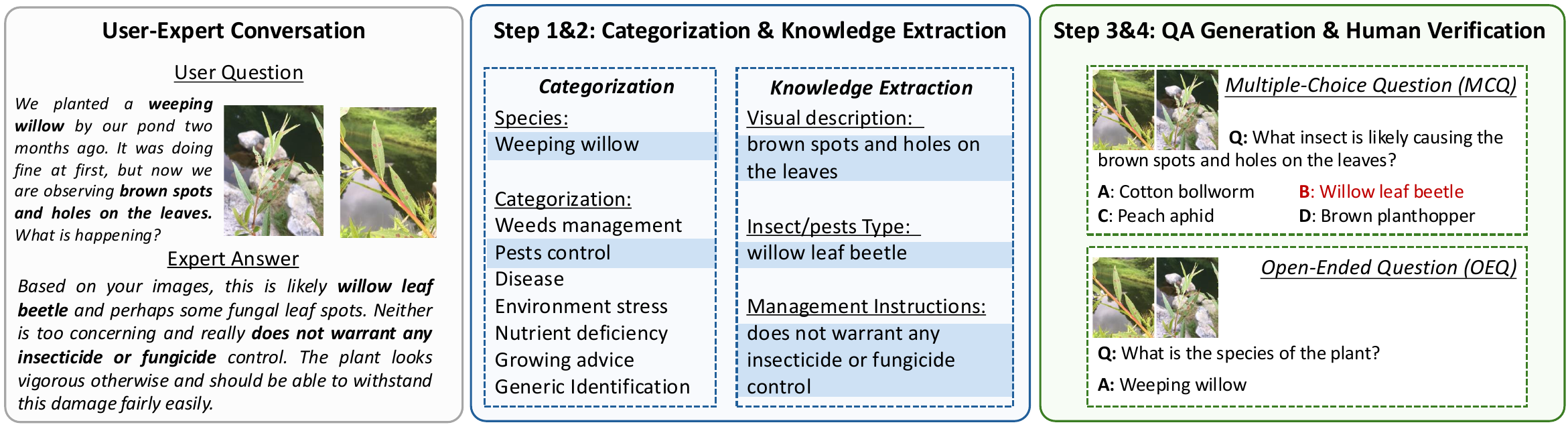}
    \vspace{-6.5mm}
    \captionof{figure}{Starting from raw user-expert conversations, we design a four-step data curation pipeline with carefully designed prompts and human verification. (1) We employ LLaMA-70B to categorize the conversation and filter out the samples that do not fall under our selected \textit{agriculture sub-domains}. (2) A larger LLaMA-405B model extracts key agricultural knowledge from the long-form user-expert conversation. (3) These facts either go to the evaluation set or the development set. For evaluation questions, we utilize GPT4o to format the original QA and agricultural knowledge into multiple-choice questions (MCQs) and open-ended questions (OEQs). (4) Finally, human annotators verify the quality of the questions and only keep the qualified ones in the evaluation set. }
    \label{fig:dataset_curation}
    \vspace{-6mm}
\end{figure*}

\vspace{-3mm}
\mypar{Curation Pipeline.} Each real-world conversation is a multi-turn QA between the user and expert, containing one or several pieces of factual agricultural knowledge. We design a four-step pipeline to convert such a conversation into one QA pair suitable for the VLM evaluation as in Fig.~\ref{fig:dataset_curation}. \textbf{(1) Categorization.} We first tag user-expert QAs with an agriculture domain and extract the primary plant organism of interest. \textbf{(2) Knowledge Extraction.} Based on the domain tag, we further decompose each expert-user conversation into its most important pieces of agricultural knowledge, where our prompts vary for different domains tagged in the previous step. \textbf{(3) QA Generation.} This step is for the evaluation set. We convert extracted agricultural facts into multiple-choice questions (MCQs) or open-ended questions (OEQs) for VLM evaluation. \textbf{(4)} \textbf{Human Verification.} We hire human annotators to filter out problematic QAs in the evaluation set.

\mypar{Step 1: Categorization.} Real-world agricultural questions come in a great variety with an open-ended nature. To extract clear agricultural knowledge, we must first account for the inherent complexity of these questions. Therefore, we first tag each question with an \textit{agriculture sub-domain}. The seven categories we decide on are $[$``\textit{disease advice}'', ``\textit{weed management}'', ``\textit{pests control}'', ``\textit{growing advice}'', ``\textit{environmental stress}'', ``\textit{nutrient deficiency}'', ``\textit{generic identification"}, and \textit{other}'$]$, following the main categories designed by gardening and agriculture sites~\cite{yardandgarden, wischorticulture, plantvillage,MESHRAM2021100010,springerReviewKnowledge}. In this step, we filter out conversations in the ``growing advice'' and ``other'' sub-domains, as these are typically not image-related questions. Notably, the type of knowledge in user questions and expert answers often depends significantly on the question's sub-domain, which we use to guide the next step, where prompts are selected according to the specific agricultural sub-domain.

\mypar{Step 2: Knowledge Extraction.} We extract agricultural knowledge for training and evaluating VLMs without introducing redundant languages from the original conversations, as in Fig.~\ref{fig:dataset_curation}. Across all questions, the user-expert conversation generally comprises a combination of identification (\textit{e.g.}, disease, or plant species) and management instructions. For example, in Fig.~\ref{fig:dataset_curation}, the user provides species identification information and symptom description, while the expert recognizes the diseases and offers management suggestions. To systematically extract knowledge, we organize the agricultural knowledge into categories of ``\textit{species identification}’’,  ``\textit{disease identification}'', ``\textit{symptom/visual description}'', ``\textit{management instructions}'', and ``\textit{insect/pest identification}.'' 

To increase the reliability of information extraction, we leverage the agriculture sub-domain predicted in the previous step and design different prompt templates for each sub-domain. Our prompt leverages the following techniques: (1) Manually annotated in-context examples, which significantly improve knowledge extraction quality and avoiding hallucination. (2) Explicit fact verification, which determines whether the expert has provided relevant facts for the target category. We discover that LLMs~\cite{dubey2024llama3,gpt4v} can reliably complete the information extraction step. More details of our prompts and observations are available in the supplementary materials.
\mypar{Step 3: QA Generation.} For rigorous evaluation, \ourwork adopts multiple-choice questions (MCQs) and open-ended questions (OEQs), combining the advantages of MMMU~\cite{yue2024mmmu} and SimpleQA~\cite{wei2024simpleqa}, respectively. For the knowledge extracted from the previous step, a small and balanced subset is selected for evaluation, while the rest goes directly to \textbf{\textsc{AgBase}} for model development and fine-tuning. We use GPT-4o~\cite{gpt4v} to format agricultural facts into QA pairs. 

The following criteria guide the selection of the evaluation set via our close observation of the conversations. (1) Filtering is done by removing the conversations \emph{without} management information or issue identification, as we empirically find that the absence of such information indicates expert uncertainty. We also filter out questions with uncertain terms, such as ``unclear'' in the extracted information. (2) We remove questions with management instruction terms that direct users to an external source rather than giving them an actual answer, typically containing the words ``lab,'' ``arborist,'' etc. (3) To balance between different \textit{agriculture sub-domains} (step 1) and \textit{knowledge types} (step 2), we prioritize an even distribution across the types of questions, while maximizing the diversity of categories, as shown in Fig.~\ref{fig:figure3}. 
%This process extracts 8,000 facts with high-quality and balanced distribution for evaluation. 

Then we employ GPT4o~\cite{gpt4v} to convert these facts into MCQs and OEQs, without losing the original intention of the user's question. Depending on the \textit{knowledge types} of the fact, we give the LLM a unique prompt to construct a corresponding question. We prompt it to generate three wrong answers and use the extracted fact as the ground truth to form one correct answer. 
%This process gives us 5,460 unique MCQs and OEQs paired with 12,481 unique images. 
%To ensure \textit{fairness} of the question, we extract ``\textit{background information}'' with LLaMA-405B from the original question which contains relevant information that the user provided. We also include the month,and location information in the question. 

\mypar{Step 4: Human verification} To guarantee the quality of the evaluation questions, we hire human annotators to manually verify the QA pairs from the following perspectives: (1) \textbf{Faithfulness}: whether the QA pair is faithful to the original conversation; (2) \textbf{Certainty}: whether the expert is certain about the answer; (3) \textbf{Quality}: are the images of good quality and correspond to the symptoms described by the user and the expert; (4) \textbf{MCQ Feasibility}: are all options other than the ground truth wrong. We strictly keep the data that pass all the evaluation criteria, resulting in 746 highest-quality MCQ and OEQ questions in \textbf{\ourwork} from 1742 samples.

\mypar{Ethical Compliance.} We use automatic procedures assisted with human verification to erase the information that could leak the name, gender, address, or any other personal information for both users and experts. We will only distribute the processed \ourwork instead of the source data from the original conversation to protect the privacy of the users and experts. To the best of our verification, we have not found any human faces in the benchmark images.

\vspace{-2mm}
\subsection{Additional Properties of \ourwork}
\vspace{-2mm}

\begin{wrapfigure}{r}{0.54\textwidth}
        \centering
        \vspace{-6mm}
        \includegraphics[width=\linewidth]{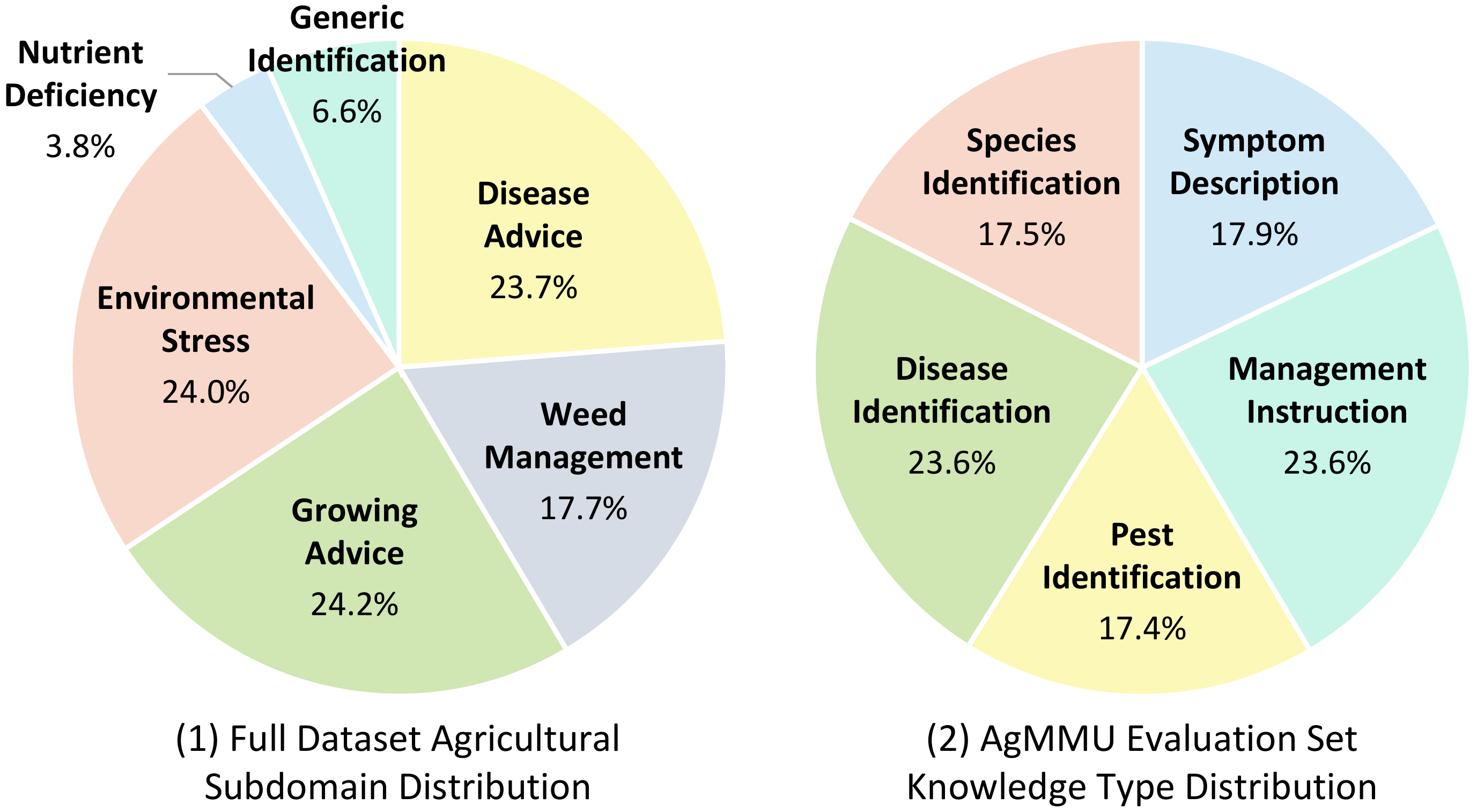}\vspace{-2mm}
        \captionsetup{font=footnotesize}
        \caption{\textbf{Statistics of \ourwork}. (1) The agricultural sub-domain distribution of our raw dataset, after the categorization step, as explained in Figure~\ref{fig:dataset_curation}. (2) \ourwork, after the knowledge extraction and evaluation curation steps, serves as a balanced subset of raw dataset with proportional representation across knowledge types.}
        \vspace{-4mm}
        \label{fig:figure3}
\end{wrapfigure}
\textbf{Distribution and Coverage.} In Fig.~\ref{fig:figure3}, we show the distribution of agricultural subdomains and knowledge types in \ourwork. We have two key observations: (1) The original conversations we obtained are severely skewed in subdomains: Subdomains like nutrient deficiency and generic identification domains have significantly fewer samples than dominating domains such as disease advice or growing advice. This imbalance is inevitably carried over to knowledge types after the knowledge extraction stage. (2) However, in \ourwork, we demonstrate a much more balanced distribution with our dataset curation pipeline. This is exceptionally important for our evaluation set, where our aim is to propose an all-encompassing benchmark that examines different aspects of models without inductive bias.
\vspace{-1mm}
\mypar{Realistic Images.} Our \ourwork also features significant challenges in the quality and number of images. The images in our benchmark are generally uploaded by users in various qualities, resolutions, and aspect ratios. This differs from the photography-level images in datasets like iNaturalist~\cite{inat21} or the textbook or web-document figures like manually curated CROP~\cite{zhang2024empowering} and CDDM~\cite{liu2024multimodal}. As shown in Fig.~\ref{fig:teaser}, \ourwork realistically reflect the multimodal everyday scenarios encountered by gardeners and farmers. Our benchmark also exhibits the necessity of \emph{multi-image understanding} challenge, since the user uploaded several images and the experts rely on a joint reasoning of them to make the final conclusion.

\vspace{-3mm}
\section{Experiments}
\label{sec:analysis}
\vspace{-3mm}
We perform comprehensive evaluations on diverse vision-language models (VLMs), including both proprietary (closed-source) and open-sourced models. Our evaluation primarily focuses on zero-shot performance, utilizing either publicly available APIs or author-provided checkpoints to reflect the inherent capabilities of each model without task-specific training. Furthermore, we present fine-tuning results on LLaVA-1.5 to demonstrate the importance of our development set \textbf{\textsc{AgBase}} for optimizing performance on agriculture tasks, and highlight some unique observations. All experiments are conducted using NVIDIA A6000 GPUs, and further details regarding model configurations, API choices, and implementation details are available in the supplementary material.

\vspace{-3mm}
\subsection{Baselines}
\vspace{-3mm}
We include a broad range of state-of-the-art VLMs to ensure robust comparisons. For open-sourced models, we prioritize architectures with comparable parameter numbers for a fair and meaningful comparison. Moreover, we include the most recent, largest, and highest-performing checkpoints up to the date of our evaluation questions, including LLaMA-3.2~\cite{dubey2024llama3}, LLaVA-1.5~\cite{llava}, LLaVA-NeXT~\cite{llavanext}, LLaVA-OneVision~\cite{llavaonevision}, Cambrian-1~\cite{cambrian1}, InternVL-2~\cite{internvl2}, Qwen-VL~\cite{Qwen-VL}, and VILA~\cite{vila}, as well as proprietary models such as GPT-o4-mini~\cite{gpt4v}, Claude-3.5~\cite{claude}, and Gemini1.5-Pro~\cite{gemini}. Detailed configurations and model settings are provided in the supplementary material.

\vspace{-3mm}
\subsection{Evaluation Metrics}
\vspace{-3mm}
We evaluate the models on multiple-choice questions (MCQs) and open-ended questions (OEQs). To ensure a robust and fair evaluation, we have developed a systematic approach to minimize potential variations in model responses due to intermediate generations or formatting inconsistencies.

\noindent\textbf{For MCQs}, we report accuracy as the primary evaluation metric. We score model response with string pre-processing and matching the predicted letter with the ground truth letter. 

\noindent\textbf{For OEQs}, we implement the LLM-as-judge methodology using GPT-4.1 to grade the answers. The prompts and style are initialized from SimpleQA~\cite{wei2024simpleqa} setting, but adapted for \ourwork scenarios. The grading is divided into two categories: (1) short-form responses and (2) multi-statement responses. The \textbf{short-form responses} correspond to the questions from \textit{pest identification}, \textit{disease identification}, and \textit{species identification}, normally containing several words, and are evaluated directly against the correct answer. We assign the grades of ``correct'', ``incorrect'', ``partially correct'', and ``not attempted,'' then use the harmonic mean to calculate the final grade of ``$\text{Correct}/\text{Total}$'' and ``$\text{Correct}/(\text{Total}-\text{Partially Correct})$'', similar to SimpleQA. The \textbf{multi-statement responses}, on the other hand, are for \textit{management instructions} and \textit{symptom descriptions} questions. These categories usually consist of multiple unique and standalone statements: On average, management instruction responses contain 2.7 factual statements, and symptom/visual description responses contain 1.8 factual statements. Therefore, we have our LLM judge (a) divide both predicted and ground-truth responses into individual statements; and (b) grade each of them as a short-form response; then (c) normalize the grades according to the number of statements per question to calculate the final harmonic mean score as short-form responses.

\begin{table*}[!t]
    \centering
    \vspace{-4mm}
    \fontsize{4pt}{4.8pt}\selectfont
    \setlength\tabcolsep{2pt} % Default value: 6pt
    \renewcommand{\arraystretch}{1.1} % Adjusts the row height
    \resizebox{0.95\textwidth}{!}{
    \begin{tabular}{r |ccccc|c|ccccc|c}
    \hline
      & \multicolumn{6}{c|}{\textbf{AgMMU-OEQs} $\uparrow$} & \multicolumn{6}{c}{\textbf{AgMMU-MCQs} $\uparrow$} \\ \hline
      & 
      \rotatebox{90}{Disease} &
      \rotatebox{90}{Insect/Pest} &
      \rotatebox{90}{Species} & 
      \rotatebox{90}{Management ~ }  &
      \rotatebox{90}{Symptom} &
      \rotatebox{90}{Average} &
      \rotatebox{90}{Disease} &
      \rotatebox{90}{Insect/Pest} &
      \rotatebox{90}{Species} & 
      \rotatebox{90}{Management ~ }  &
      \rotatebox{90}{Symptom} &
      \rotatebox{90}{Average} \\
      \hline
      \rowcolor{gray!10}
      \multicolumn{13}{l}{\textit{Proprietary Models}} \\ \hline
      GPT-o4-mini~\cite{gpt4v} & 47.7 & 48.8 & 69.6 &  16.2 & 2.6 & 37.0 & 77.9 & 85.4 & 90.3 & 93.8 & 84.3 & 86.5  \\
      Gemini 1.5 Pro~\cite{gemini15} & 37.6 & 51.7 & 69.9 & 32.5 & 16.9 & 41.7 & 76.2 & 81.1 & 82.8 & 88.1 & 76.9 & 82.4  \\
      Claude 3 Haiku~\cite{claude} & 25.1 & 20.6 & 32.2 & 1.0 & 13.2 & 18.4 & 62.1 & 71.2 & 52.8 & 81.5 & 52.0 & 63.8  \\ \hline
      \rowcolor{gray!10}
      \multicolumn{13}{l}{\textit{SOTA Open-sourced Models}} \\ \hline
      LLaVA-1.5-7B~\cite{llava} & 4.9 & 22.3 & 30.6 & 1.4 & 5.0 & 12.8 & 61.9 & 64.6 & 67.6 & 77.3 & 71.5 & 69.0  \\
      LLaVA-NeXT-8B~\cite{llavanext} & 6.4 & 27.2 & 31.1 & 2.6 & 0.3 & 13.5 & 61.9 & 70.1 & 57.9 & 78.5 & 67.6 & 67.5  \\  
      LLaVA-OneVision-8B~\cite{llavaonevision} & 9.1 & 24.1 & 34.1 & 5.4 & 6.2 & 15.8 & 65.7 & 72.9 & 71.2 & 85.9 & 78.8 & 75.4  \\
      LLaVA-1.5-13B~\cite{llava} & 8.2 & 22.1 & 31.8 & 1.1 & 5.9 & 13.8 & 64.7 & 67.4 & 65.5 & 80.4 & 73.7 & 70.8  \\      
      Cambrian-8B~\cite{cambrian1} & 7.9 & 30.9 & 27.9 & 2.2 & 1.9 & 14.2 & 65.0 & 70.1 & 59.3 & 79.1 & 86.0 & 72.8  \\
      InternVL2-8B~\cite{internvl2} & 8.5 & 27.0 & 22.1 & 1.0 & 1.9 & 12.1 & 55.5 & 61.0 & 60.4 & 74.7 & 64.3 & 63.5  \\
      Qwen-VL-7B~\cite{Qwen-VL} & 10.2 & 25.2 & 36.1 & 4.1 & 6.5 & 16.4 & 55.5 & 61.0 & 60.4 & 74.7 & 64.3 & 63.5  \\
      VILA1.5-13B~\cite{vila} & 11.0 & 34.1 & 36.8 & 4.9 & 6.5 & 18.7 & 61.8 & 63.2 & 60.1 & 73.9 & 59.8 & 63.7  \\
      LLaMA-3.2-11B~\cite{dubey2024llama3} & 22.9 & 32.5 & 46.2 & 5.7 & 17.1 & 24.9 & 66.2 & 75.0 & 78.6 & 89.6 & 79.9 & 78.3  \\
   \hline  
   \end{tabular}
}
% \captionsetup{font={small}}
\vspace{-1.3mm}
 \caption{Performance of VLMs on our \ourwork OEQs and MCQs. Our evaluation set poses great challenges to existing large VLMs, including closed-source and open-sourced models.}
\label{tab:main_evaluation}
\vspace{-6mm}
\end{table*}

\vspace{-2mm}
\subsection{Zero-shot Evaluation Results}
\vspace{-2mm}
We present a comprehensive zero-shot comparison of various VLMs in Table~\ref{tab:main_evaluation}. More visualizations are presented in the supplementary material.

\mypar{Very Challenging for Existing Models.} \ourwork proves to be a very challenging benchmark across all models evaluated, and even the most advanced systems achieve moderate performance levels. Such challenges are notable, especially on open-ended OEQs, where models have to recall the knowledge precisely without relying on any pre-input options. These observations also reveal the necessity of improving the agricultural expertise of VLMs and mixing agricultural data for VLM training, where our \textsc{AgBase} could be helpful.
\mypar{Performance Across Question Types.} On MCQs, the models can rely on options, so it is significantly easier than OEQs. The level of $>$70\% accuracy is also on par with previous agricultural benchmarks~\cite{zhang2024empowering, liu2024multimodal} in MCQs. Therefore, we primarily advocate for the more challenging OEQ setting, which is also more aligned with a VLM in the real world. As observed in Table~\ref{tab:main_evaluation}, the \textit{management instruction} and \textit{symptom/visual description} types of questions pose more challenges, due to their multi-statement and long-response nature and the need for more reasoning. However, there is significant variation between models as to what tasks are the most difficult, suggesting that specialized agricultural knowledge and visual understanding capabilities are not uniformly distributed across model architectures. 

\begin{figure}[!t]
% \vspace{-4mm}
    \raggedright
    \begin{subfigure}[t]{0.66\textwidth}
        \centering
        \includegraphics[width=1\linewidth]{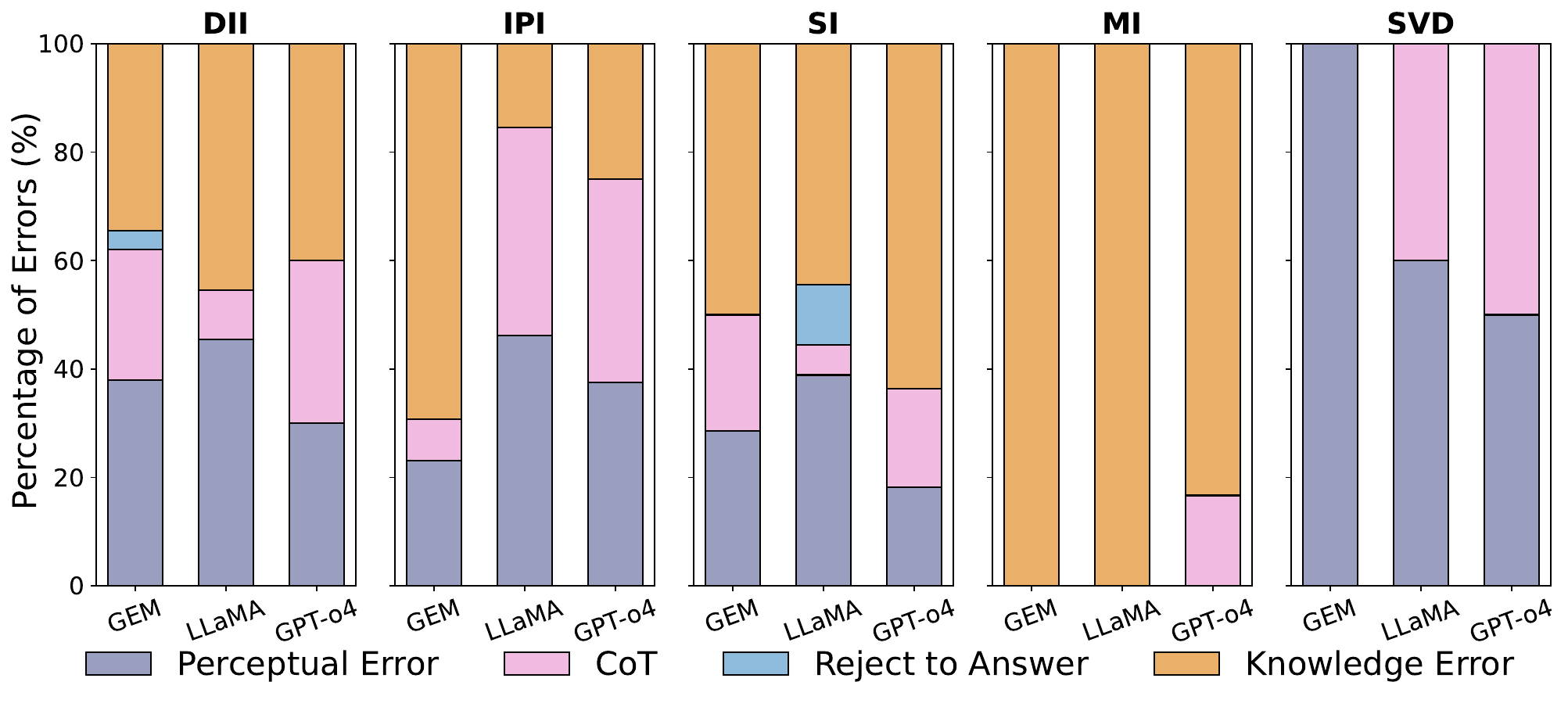}\vspace{-1mm}
        \caption{\textbf{Error Analysis}.}
        \label{fig:error-analysis}
    \end{subfigure}%
    \begin{subfigure}[t]{0.34\textwidth}
        \centering
        \includegraphics[width=1\linewidth]{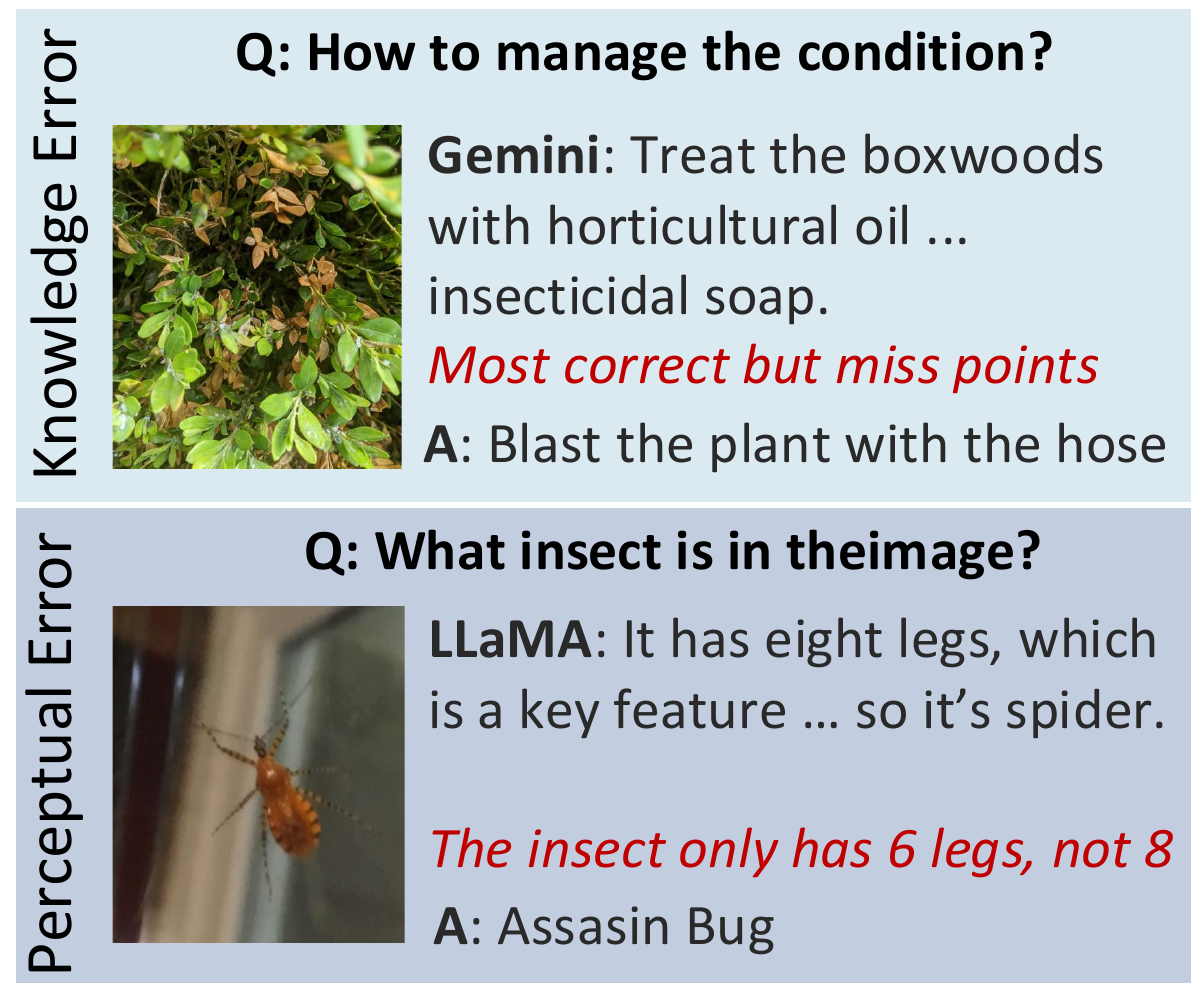}
        \vspace{-5mm}
        \caption{\textbf{Error Demonstration}.}
        \label{fig:error-analysis-figure}
    \end{subfigure}%
    \vspace{-2mm}
    \caption{(a) The most common errors made by VLMs are knowledge errors. CoT represents samples that are originally wrong due to false reasoning, but corrected with chain-of-thought prompting. IPI, DII, SVD, SI, and MI are short for five question types as explained in Sec~\ref{sec:data_curation}. (b)  We show two common evaluation errors, including the lack of knowledge to answer the question (\textit{top}), and the wrong perception of the image (\textit{bottom}).}
    \vspace{-7mm}
\end{figure}

\mypar{Closed- and Open-sourced Models.} We observe a distinct performance gap between closed-source and open-sourced models, where closed-source models generally demonstrate superior performance. Among all models, Gemini 1.5 Pro and GPT-o4-mini lead in all tasks. For open-sourced models, LLaMA-3.2 is the overall best performer across all subdomains. On average, open-sourced models trail behind their closed-source counterparts by more than 10 to 20\%, likely due to the lack of substantial agricultural data during the large-scale training phases.
\vspace{-3mm}
\subsection{Error Analysis}
\vspace{-3mm}

% \begin{wrapfigure}{r}{0.65\textwidth}
% \centering
% \vspace{-10mm}
% \includegraphics[width=0.99\linewidth]{figures/pdf/all_models_error_breakdown.pdf}
% \vspace{-3mm}
% \caption{\textbf{Error Analysis}. We categorize the errors into four representative cases. The names DII, IPI, ..., SVD are question types explained in Sec.~\ref{sec:data_curation}.}
% \label{fig:error-analysis}
% \vspace{-3mm}
% \end{wrapfigure}
To achieve an in-depth understanding of the internal procedures of VLMs in answering \ourwork questions, we analyze the types of error that occurred during the evaluation. We select 171 wrong evaluation samples and ask the VLMs to provide the chain-of-thought (CoT) reasoning before providing the final answer, and judge the error types with human evaluators. Fig.~\ref{fig:error-analysis} demonstrates quantitative results, and Fig.~\ref{fig:error-analysis-figure} shows the qualitative results of the most common knowledge and perceptual errors.

\mypar{Perceptual Error.} A perceptual error occurs when the VLM fails to recognize a primary visual characteristic or gives a direct incorrect description of a visual characteristic as in Fig.~\ref{fig:error-analysis-figure}. This category highlights limitations in the visual understanding capabilities of VLMs. Perceptual errors constitute 34.3\% (Gemini), 37.5\% (LLaMA), and 22.2\% (GPT-o4) of total errors. 

% \begin{wrapfigure}{r}{0.56\textwidth}
% \centering
% \vspace{-4mm}
% \includegraphics[width=0.99\linewidth]{figures/pdf/error_analysis_figure_v2.pdf}
% \vspace{-3mm}
% \caption{\textbf{Error Visualization}. Knowledge and perceptual errors are the most common for VLMs because of a lack of knowledge or failing to recognize the details.}\vspace{-6mm}
% \label{fig:error-analysis-figure}
% \end{wrapfigure}
\mypar{Knowledge Error.} A knowledge error 
occurs when the VLM does not have the essential knowledge to reach the correct answer. In these cases, the model may identify relevant visual features and reinforce it, but without reaching the correct answer. For example, in management questions, the model might give vague responses that do not engage with the particular nuances of the given situation (Fig.~\ref{fig:error-analysis-figure}). Knowledge errors constitute the highest proportion of errors for all Gemini (48.6\%), LLaMA (42.9\%), and GPT-o4 (51.1\%).

\mypar{Chain-of-Thought (CoT).} This category includes instances where the VLM initially provides an incorrect response but corrects itself after being asked to explain its reasoning. The proportion of CoT errors underscores the importance of reasoning processes for accurate performance on our benchmark. Specifically, CoT accounts for 15.7\% (Gemini), 16.1\% (LLaMA), and 26.7\% (GPT-o4) of total errors, indicating that these models benefit significantly from explicit reasoning steps.

\mypar{Reject to Answer.} Occasionally, VLMs refuse to answer a question entirely, reflecting uncertainty or lack of confidence in their predictions. This occurs rarely, constituting 1.4\% (Gemini) and 3.6\% (LLaMA), with no instances observed in GPT-o4.

The differing error patterns across Gemini, LLaMA, and GPT-o4 reveal distinct model behaviors. Perceptual ability forms the foundation: models like LLaMA, with weaker visual grounding, show more perceptual errors—often hallucinating features to match textual cues. In contrast, Gemini accurately identifies main visual elements without hallucination but frequently makes knowledge errors, indicating gaps in domain understanding despite solid perception. GPT-o4, with strong perception and knowledge, rarely makes blatant errors, but its reasoning steps can still misfire.

\begin{figure*}[!t]
    \raggedright
    \vspace{-4mm}
    \includegraphics[width=1\linewidth]{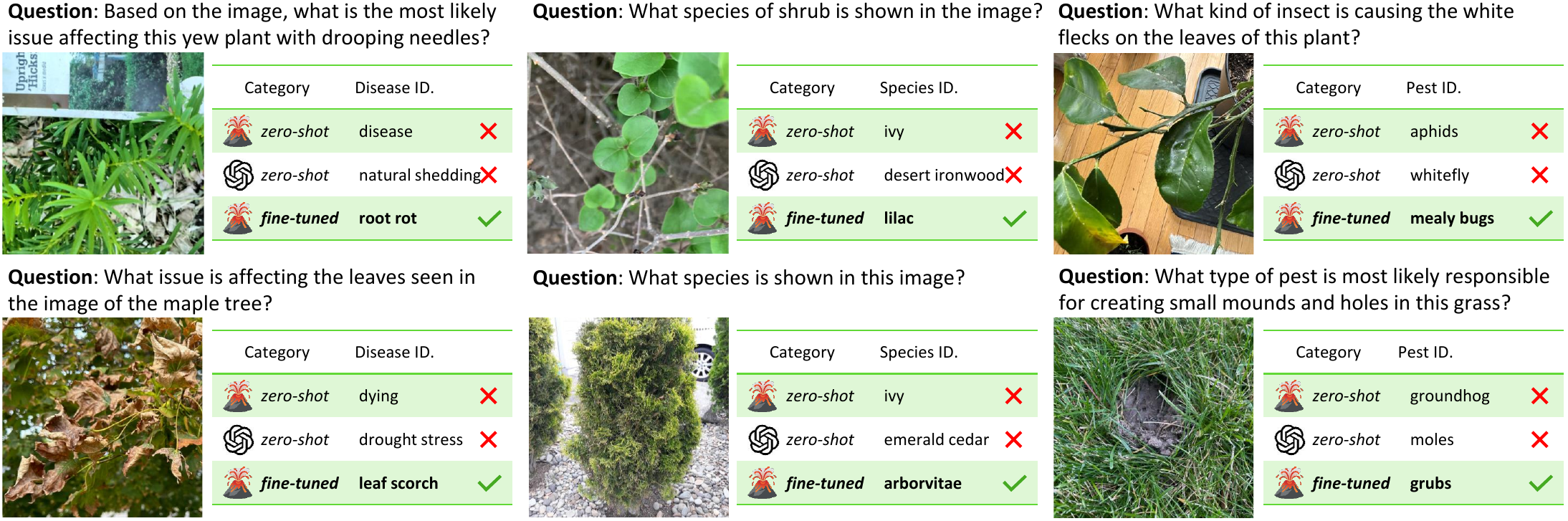}
    \vspace{-5.5mm}
    \caption{The effectiveness of \textsc{AgBase} fine-tuning on OEQ examples. After simple fine-tuning, the LLaVA model can accurately identify issues that GPT and the original model fail to recognize in zero-shot scenarios.}
    \label{fig:finetuning-mcq-vis}
    \vspace{-5mm}
\end{figure*}

\vspace{-2mm}
\subsection{Finetuning Evaluation Results}
\vspace{-2mm}

\begin{wrapfigure}{r}{0.5\textwidth}
        \centering
        \vspace{-7mm}
        \includegraphics[width=\linewidth]{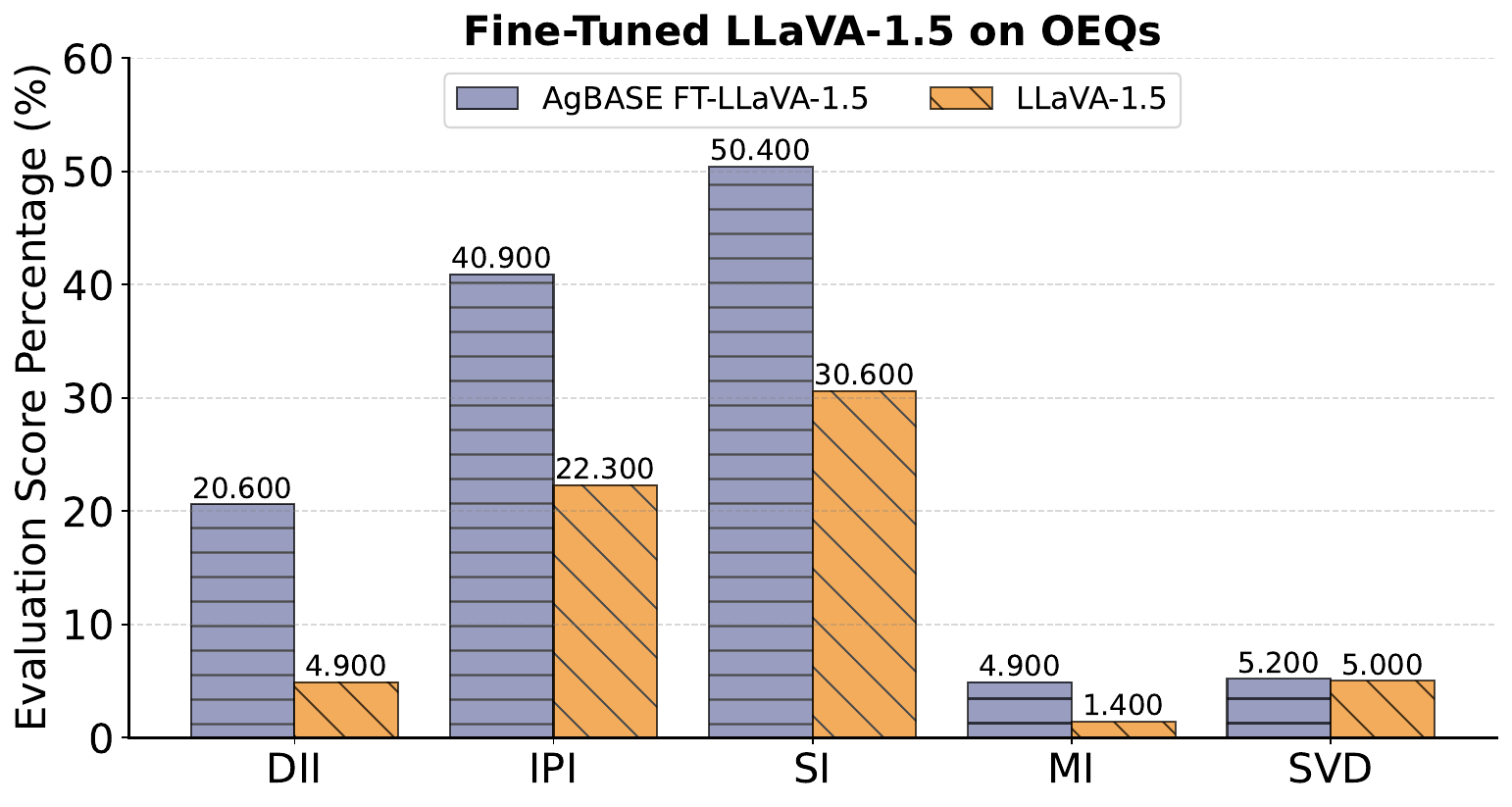}
	\vspace{-5.5mm}
        % \captionsetup{font=footnotesize}
        \caption{Fine-tuning on \textsc{AgBase} boosts the agricultural knowledge understanding for LLaVA-1.5, indicating the potential of our development set. IPI, DII, SVD, SI, and MI are the short for the five question types as explained in Sec.~\ref{sec:data_curation}.}
        \vspace{-4mm}
        \label{fig:finetuning-mcq-qtype}
\end{wrapfigure}
\textbf{Experiment Setting.} We finetune LLaVA-1.5-7B~\cite{llava} using our \textsc{AgBase}. Our training follows the standard practice of LLaVA by combining its instruction-tuning set and \textsc{AgBase} together, where the instruction-tuning set of LLaVA is critical for instruction-following abilities to correctly answer MCQs. Our training lasts two epochs using LoRA adapters~\cite{hu2021lora}, a learning rate of 2e-4, a weight decay of 0, and a batch size 16. The fine-tuning process takes approximately 12 hours on 2$\times$A6000 GPUs. 

\vspace{-1mm}
\mypar{Analysis.} Fig.~\ref{fig:finetuning-mcq-vis} and Fig.~\ref{fig:finetuning-mcq-qtype} demonstrate the effectiveness of fine-tuning VLMs using our \textsc{AgBase} on the more challenging open-ended questions. Fig.~\ref{fig:finetuning-mcq-qtype} shows that fine-tuning with our knowledge base significantly improves the model's capability of understanding the images and correctly responding with agriculture knowledge, leading to up to 11.6\% improvement on average. This performance boost highlights the effectiveness of our large-scale knowledge base in improving VLMs and the necessity of collecting agricultural-related data for future VLMs. Meanwhile, the smaller improvement on management instruction and symptom description questions indicates the complexity of multi-statement questions, motivating the need for more data collection and better model training than simple fine-tuning.

\vspace{-3mm}
\section{Conclusion}
\label{sec:conclusions}
\vspace{-3mm}
In this work, we introduced \ourwork, the first benchmark for evaluating vision-language models (VLMs) in agriculture, a field that requires precise visual interpretation and expert knowledge. Our dataset spans core agricultural tasks, including symptom recognition, species and pest identification, and management instructions, and is created from more than 116,231 real-world expert-user interactions. Through a three-step curation framework, we build a high-quality \ourwork evaluation set, demonstrating a great challenge for current VLMs. To support model development, we introduce \textsc{AgBase}, a training set of 57,079 knowledge entries aimed at improving model accuracy. We hope that the combination of \textsc{AgMMU} and \textsc{AgBase} can support the community in evaluating and developing stronger knowledge-intensive VLMs.

% \paragraph{Acknowledgments.}
% We thank Kastan Day and Rohan Marwaha for the fruitful discussion. We thank the \href{https://ask.extension.org/}{AskExtension} team for helping us with the data curation. This work was supported in part by the AIFARMS National AI Institute, USDA-NIFA Award 2020-67021-32799, NSF Grant 2106825, the University of Illinois Discovery Partners Institute, the Center for Digital Agriculture at the University of Illinois, the Amazon-Illinois Center on AI for Interactive Conversational Experiences, and the IBM-Illinois Discovery Accelerator Institute. This work used computational resources, including the NCSA Delta and DeltaAI supercomputers through allocations CIS230012 and CIS240133 from the Advanced Cyberinfrastructure Coordination Ecosystem: Services \& Support (ACCESS) program, as well as the TACC Frontera supercomputer, Amazon Web Services (AWS), and OpenAI API through the National Artificial Intelligence Research Resource (NAIRR) Pilot.

{
    \small
    \bibliographystyle{ieeenat_fullname}
    \bibliography{main}

\begin{thebibliography}{57}
\providecommand{\natexlab}[1]{#1}
\providecommand{\url}[1]{\texttt{#1}}
\expandafter\ifx\csname urlstyle\endcsname\relax
  \providecommand{\doi}[1]{doi: #1}\else
  \providecommand{\doi}{doi: \begingroup \urlstyle{rm}\Url}\fi

\bibitem[Achiam et~al.(2023)Achiam, Adler, Agarwal, Ahmad, Akkaya, Aleman, Almeida, Altenschmidt, Altman, Anadkat, et~al.]{gpt4v}
Josh Achiam, Steven Adler, Sandhini Agarwal, Lama Ahmad, Ilge Akkaya, Florencia~Leoni Aleman, Diogo Almeida, Janko Altenschmidt, Sam Altman, Shyamal Anadkat, et~al.
\newblock {GPT-4} technical report.
\newblock \emph{arXiv preprint arXiv:2303.08774}, 2023.

\bibitem[Adve et~al.(2024)Adve, Day, Dabholkar, and Marwaha]{CropWizard:URL}
V Adve, K Day, A Dabholkar, and R Marwaha.
\newblock Cropwizard: Visual and computational question answering for agriculture professionals.
\newblock \url{https://uiuc.chat/cropwizard-1.5/}, 2024.

\bibitem[Awais et~al.(2024)Awais, Alharthi, Kumar, Cholakkal, and Anwer]{awais2024agrogptefficientagriculturalvisionlanguage}
Muhammad Awais, Ali Husain Salem~Abdulla Alharthi, Amandeep Kumar, Hisham Cholakkal, and Rao~Muhammad Anwer.
\newblock Agrogpt: Efficient agricultural vision-language model with expert tuning, 2024.

\bibitem[Bai et~al.(2023)Bai, Bai, Yang, Wang, Tan, Wang, Lin, Zhou, and Zhou]{Qwen-VL}
Jinze Bai, Shuai Bai, Shusheng Yang, Shijie Wang, Sinan Tan, Peng Wang, Junyang Lin, Chang Zhou, and Jingren Zhou.
\newblock Qwen-vl: A versatile vision-language model for understanding, localization, text reading, and beyond.
\newblock \emph{arXiv preprint arXiv:2308.12966}, 2023.

\bibitem[Balaguer et~al.(2024{\natexlab{a}})Balaguer, Benara, Cunha, Hendry, Holstein, Marsman, Mecklenburg, Malvar, Nunes, Padilha, et~al.]{balaguer2024rag}
Angels Balaguer, Vinamra Benara, Renato Luiz de~Freitas Cunha, Todd Hendry, Daniel Holstein, Jennifer Marsman, Nick Mecklenburg, Sara Malvar, Leonardo~O Nunes, Rafael Padilha, et~al.
\newblock Rag vs fine-tuning: Pipelines, tradeoffs, and a case study on agriculture.
\newblock \emph{arXiv preprint arXiv:2401.08406}, 2024{\natexlab{a}}.

\bibitem[Balaguer et~al.(2024{\natexlab{b}})Balaguer, Benara, de~Freitas~Cunha, de~M.~Estevão~Filho, Hendry, Holstein, Marsman, Mecklenburg, Malvar, Nunes, Padilha, Sharp, Silva, Sharma, Aski, and Chandra]{balaguer2024ragvsfinetuningpipelines}
Angels Balaguer, Vinamra Benara, Renato~Luiz de Freitas~Cunha, Roberto de M.~Estevão~Filho, Todd Hendry, Daniel Holstein, Jennifer Marsman, Nick Mecklenburg, Sara Malvar, Leonardo~O. Nunes, Rafael Padilha, Morris Sharp, Bruno Silva, Swati Sharma, Vijay Aski, and Ranveer Chandra.
\newblock Rag vs fine-tuning: Pipelines, tradeoffs, and a case study on agriculture, 2024{\natexlab{b}}.

\bibitem[Bayer Crop~Science(2024)]{BayerAITool:URL}
Inc. Bayer Crop~Science.
\newblock Bayer pilots unique generative ai tool for agriculture.
\newblock \url{ https://www.bayer.com/media/en-us/bayer-pilots-unique-generative-ai-tool-for-agriculture/}, 2024.

\bibitem[Chiranjeevi et~al.(2023)Chiranjeevi, Sadaati, ZK, Koushik, TZ, Mueller, O'Neal, Merchant, Singh, Singh, Sarkar, Singh, and Ganapathysubramanian]{chiranjeevi:arXiv23}
S Chiranjeevi, M Sadaati, Deng ZK, J Koushik, Jubery TZ, D Mueller, ME O'Neal, N Merchant, A Singh, AK Singh, S Sarkar, A Singh, and B Ganapathysubramanian.
\newblock Deep learning powered real-time identification of insects using citizen science data.
\newblock \emph{arXiv preprint arXiv:2306.02507}, 2023.

\bibitem[Cieslak et~al.(2024)Cieslak, Govindarajan, Garcia, Chandrashekar, Hadrich, Mendoza-Drosik, Michels, Pirk, Fu, and Palubicki]{cieslak2024generating}
Mikolaj Cieslak, Umabharathi Govindarajan, Alejandro Garcia, Anuradha Chandrashekar, Torsten Hadrich, Aleksander Mendoza-Drosik, Dominik~L Michels, Soren Pirk, Chia-Chun Fu, and Wojciech Palubicki.
\newblock Generating diverse agricultural data for vision-based farming applications.
\newblock In \emph{CVPRW}, 2024.

\bibitem[Digital~Green(2023)]{DigitalGreen:Farmer.Chat:URL}
Inc. Digital~Green.
\newblock Farmer.chat by digital green – making vetted farmer knowledge accessible.
\newblock \url{ https://digitalgreen.org/farmerchat/}, 2023.

\bibitem[Dubey et~al.(2024)Dubey, Jauhri, Pandey, Kadian, Al-Dahle, Letman, Mathur, Schelten, Yang, Fan, et~al.]{dubey2024llama3}
Abhimanyu Dubey, Abhinav Jauhri, Abhinav Pandey, Abhishek Kadian, Ahmad Al-Dahle, Aiesha Letman, Akhil Mathur, Alan Schelten, Amy Yang, Angela Fan, et~al.
\newblock The llama 3 herd of models.
\newblock \emph{arXiv preprint arXiv:2407.21783}, 2024.

\bibitem[(FBN)(2023)]{FBNNorm:URL}
The Farmers Business~Network (FBN).
\newblock Meet norm, the world’s first ai ag advisor.
\newblock \url{https://www.fbn.com/norm}, 2023.

\bibitem[Fenu and Malloci(2021)]{agronomy11112107}
Gianni Fenu and Francesca~Maridina Malloci.
\newblock Diamos plant: A dataset for diagnosis and monitoring plant disease.
\newblock \emph{Agronomy}, 11\penalty0 (11), 2021.

\bibitem[Feuer et~al.(2024)Feuer, Joshi, Cho, Chiranjeevi, Deng, Balu, Singh, Sarkar, Merchant, and Singh]{InsectDataSet:2024}
Benjamin Feuer, Ameya Joshi, Minsu Cho, Shivani Chiranjeevi, Zi~Kang Deng, Aditya Balu, Asheesh~K. Singh, Soumik Sarkar, Nirav Merchant, and Arti Singh.
\newblock Zero-shot insect detection via weak language supervision.
\newblock \emph{The Plant Phenome Journal}, 7, 2024.

\bibitem[Foundation(2023)]{extensionforum}
Extention Foundation.
\newblock Ask extension.
\newblock \url{https://extension.org/tools/ask-extension/}, 2023.

\bibitem[Gharaee et~al.(2024)Gharaee, Gong, Pellegrino, Zarubiieva, Haurum, Lowe, McKeown, Ho, McLeod, Wei, et~al.]{gharaee2024bioscan}
Zahra Gharaee, ZeMing Gong, Nicholas Pellegrino, Iuliia Zarubiieva, Joakim~Bruslund Haurum, Scott Lowe, Jaclyn McKeown, Chris Ho, Joschka McLeod, Yi-Yun Wei, et~al.
\newblock A step towards worldwide biodiversity assessment: The bioscan-1m insect dataset.
\newblock \emph{NeurIPS}, 2024.

\bibitem[Gong et~al.(2024)Gong, Wang, Haurum, Lowe, Taylor, and Chang]{gong2024bioscan}
ZeMing Gong, Austin~T Wang, Joakim~Bruslund Haurum, Scott~C Lowe, Graham~W Taylor, and Angel~X Chang.
\newblock Bioscan-clip: Bridging vision and genomics for biodiversity monitoring at scale.
\newblock \emph{arXiv preprint arXiv:2405.17537}, 2024.

\bibitem[Goyal et~al.(2017)Goyal, Khot, Summers{-}Stay, Batra, and Parikh]{balanced_vqa_v2}
Yash Goyal, Tejas Khot, Douglas Summers{-}Stay, Dhruv Batra, and Devi Parikh.
\newblock Making the {V} in {VQA} matter: Elevating the role of image understanding in {V}isual {Q}uestion {A}nswering.
\newblock In \emph{CVPR}, 2017.

\bibitem[Horn and Aodha(2021)]{inat21}
Grant~Van Horn and Oisin~Mac Aodha.
\newblock inat challenge 2021, 2021.

\bibitem[Hu et~al.(2021)Hu, Wallis, Allen-Zhu, Li, Wang, Wang, Chen, et~al.]{hu2021lora}
Edward~J Hu, Phillip Wallis, Zeyuan Allen-Zhu, Yuanzhi Li, Shean Wang, Lu Wang, Weizhu Chen, et~al.
\newblock Lora: Low-rank adaptation of large language models.
\newblock In \emph{ICLR}, 2021.

\bibitem[Hudson and Manning(2019)]{hudson2019gqa}
Drew~A Hudson and Christopher~D Manning.
\newblock Gqa: A new dataset for real-world visual reasoning and compositional question answering.
\newblock In \emph{CVPR}, 2019.

\bibitem[Li et~al.(2024)Li, Zhang, Guo, Zhang, Li, Zhang, Zhang, Li, Liu, and Li]{llavaonevision}
Bo Li, Yuanhan Zhang, Dong Guo, Renrui Zhang, Feng Li, Hao Zhang, Kaichen Zhang, Yanwei Li, Ziwei Liu, and Chunyuan Li.
\newblock {LLaVA-OneVision}: Easy visual task transfer.
\newblock \emph{arXiv preprint arXiv:2408.03326}, 2024.

\bibitem[Li et~al.(2023)Li, Wong, Zhang, Usuyama, Liu, Yang, Naumann, Poon, and Gao]{li2023llavamedtraininglargelanguageandvision}
Chunyuan Li, Cliff Wong, Sheng Zhang, Naoto Usuyama, Haotian Liu, Jianwei Yang, Tristan Naumann, Hoifung Poon, and Jianfeng Gao.
\newblock Llava-med: Training a large language-and-vision assistant for biomedicine in one day, 2023.

\bibitem[Lin et~al.(2024)Lin, Yin, Ping, Molchanov, Shoeybi, and Han]{vila}
Ji Lin, Hongxu Yin, Wei Ping, Pavlo Molchanov, Mohammad Shoeybi, and Song Han.
\newblock {VILA}: On pre-training for visual language models.
\newblock In \emph{CVPR}, 2024.

\bibitem[Liu et~al.(2023)Liu, Li, Wu, and Lee]{llava}
Haotian Liu, Chunyuan Li, Qingyang Wu, and Yong~Jae Lee.
\newblock Visual instruction tuning.
\newblock In \emph{NeurIPS}, 2023.

\bibitem[Liu et~al.(2024{\natexlab{a}})Liu, Li, Li, Li, Zhang, Shen, and Lee]{llavanext}
Haotian Liu, Chunyuan Li, Yuheng Li, Bo Li, Yuanhan Zhang, Sheng Shen, and Yong~Jae Lee.
\newblock {LLaVA-NeXT}: Improved reasoning, {OCR}, and world knowledge, 2024{\natexlab{a}}.

\bibitem[Liu et~al.(2021)Liu, Min, Mei, Wang, and Jiang]{9325065}
Xinda Liu, Weiqing Min, Shuhuan Mei, Lili Wang, and Shuqiang Jiang.
\newblock Plant disease recognition: A large-scale benchmark dataset and a visual region and loss reweighting approach.
\newblock \emph{IEEE Transactions on Image Processing}, 30:\penalty0 2003--2015, 2021.

\bibitem[Liu et~al.(2024{\natexlab{b}})Liu, Liu, Hu, Chen, Wang, Wang, and Lian]{liu2024multimodal}
Xiang Liu, Zhaoxiang Liu, Huan Hu, Zezhou Chen, Kohou Wang, Kai Wang, and Shiguo Lian.
\newblock A multimodal benchmark dataset and model for crop disease diagnosis.
\newblock In \emph{ECCV}, 2024{\natexlab{b}}.

\bibitem[Maruf et~al.(2024)Maruf, Daw, Mehrab, Manogaran, Neog, Sawhney, Khurana, Balhoff, Bakis, Altintas, et~al.]{maruf2024vlm4bio}
M Maruf, Arka Daw, Kazi~Sajeed Mehrab, Harish~Babu Manogaran, Abhilash Neog, Medha Sawhney, Mridul Khurana, James~P Balhoff, Yasin Bakis, Bahadir Altintas, et~al.
\newblock Vlm4bio: A benchmark dataset to evaluate pretrained vision-language models for trait discovery from biological images.
\newblock \emph{arXiv preprint arXiv:2408.16176}, 2024.

\bibitem[Mathew et~al.(2021)Mathew, Karatzas, and Jawahar]{mathew2021docvqa}
Minesh Mathew, Dimosthenis Karatzas, and CV Jawahar.
\newblock Docvqa: A dataset for vqa on document images.
\newblock In \emph{WACV}, 2021.

\bibitem[Meshram et~al.(2021)Meshram, Patil, Meshram, Hanchate, and Ramkteke]{MESHRAM2021100010}
Vishal Meshram, Kailas Patil, Vidula Meshram, Dinesh Hanchate, and S.D. Ramkteke.
\newblock Machine learning in agriculture domain: A state-of-art survey.
\newblock \emph{Artificial Intelligence in the Life Sciences}, 1:\penalty0 100010, 2021.

\bibitem[Nismi Mol~E.A.(2022)]{springerReviewKnowledge}
M.B. Nismi Mol~E.A., Santosh~Kumar.
\newblock {R}eview on knowledge extraction from text and scope in agriculture domain - {A}rtificial {I}ntelligence {R}eview --- link.springer.com, 2022.

\bibitem[Patrício and Rieder(2018)]{PATRICIO201869}
Diego~Inácio Patrício and Rafael Rieder.
\newblock Computer vision and artificial intelligence in precision agriculture for grain crops: A systematic review.
\newblock \emph{Computers and Electronics in Agriculture}, 153:\penalty0 69--81, 2018.

\bibitem[Reid et~al.(2024)Reid, Savinov, Teplyashin, Lepikhin, Lillicrap, Alayrac, Soricut, Lazaridou, Firat, Schrittwieser, et~al.]{gemini15}
Machel Reid, Nikolay Savinov, Denis Teplyashin, Dmitry Lepikhin, Timothy Lillicrap, Jean-baptiste Alayrac, Radu Soricut, Angeliki Lazaridou, Orhan Firat, Julian Schrittwieser, et~al.
\newblock {Gemini 1.5}: Unlocking multimodal understanding across millions of tokens of context.
\newblock \emph{arXiv preprint arXiv:2403.05530}, 2024.

\bibitem[Saikh et~al.(2022)Saikh, Ghosal, Mittal, Ekbal, and Bhattacharyya]{saikh2022scienceqa}
Tanik Saikh, Tirthankar Ghosal, Amish Mittal, Asif Ekbal, and Pushpak Bhattacharyya.
\newblock Scienceqa: A novel resource for question answering on scholarly articles.
\newblock \emph{IJDL}, 2022.

\bibitem[Silva et~al.(2023)Silva, Nunes, Estev{\~a}o, Aski, and Chandra]{silva2023gpt}
Bruno Silva, Leonardo Nunes, Roberto Estev{\~a}o, Vijay Aski, and Ranveer Chandra.
\newblock Gpt-4 as an agronomist assistant? answering agriculture exams using large language models.
\newblock \emph{arXiv preprint arXiv:2310.06225}, 2023.

\bibitem[Singh et~al.(2019)Singh, Natarajan, Shah, Jiang, Chen, Batra, Parikh, and Rohrbach]{textvqa}
Amanpreet Singh, Vivek Natarajan, Meet Shah, Yu Jiang, Xinlei Chen, Dhruv Batra, Devi Parikh, and Marcus Rohrbach.
\newblock Towards {VQA} models that can read.
\newblock In \emph{CVPR}, 2019.

\bibitem[Singh et~al.(2020)Singh, Jain, Jain, Kayal, Kumawat, and Batra]{10.1145/3371158.3371196}
Davinder Singh, Naman Jain, Pranjali Jain, Pratik Kayal, Sudhakar Kumawat, and Nipun Batra.
\newblock Plantdoc: A dataset for visual plant disease detection.
\newblock In \emph{Proceedings of the 7th ACM IKDD CoDS and 25th COMAD}, page 249–253, New York, NY, USA, 2020. Association for Computing Machinery.

\bibitem[Stevens et~al.(2024)Stevens, Wu, Thompson, Campolongo, Song, Carlyn, Dong, Dahdul, Stewart, Berger-Wolf, et~al.]{stevens2024bioclip}
Samuel Stevens, Jiaman Wu, Matthew~J Thompson, Elizabeth~G Campolongo, Chan~Hee Song, David~Edward Carlyn, Li Dong, Wasila~M Dahdul, Charles Stewart, Tanya Berger-Wolf, et~al.
\newblock Bioclip: A vision foundation model for the tree of life.
\newblock In \emph{CVPR}, 2024.

\bibitem[Taranis(2024)]{Taranis:URL}
Inc. Taranis.
\newblock Taranis launches ag assistant to transform farm decision making.
\newblock \url{https://www.taranis.com/newsroom/ag-assistant/}, 2024.

\bibitem[Team(2024{\natexlab{a}})]{claude}
Anthropic Team.
\newblock Introducing the next generation of claude, 2024{\natexlab{a}}.

\bibitem[Team et~al.(2023)Team, Anil, Borgeaud, Wu, Alayrac, Yu, Soricut, Schalkwyk, Dai, Hauth, et~al.]{gemini}
Gemini Team, Rohan Anil, Sebastian Borgeaud, Yonghui Wu, Jean-Baptiste Alayrac, Jiahui Yu, Radu Soricut, Johan Schalkwyk, Andrew~M Dai, Anja Hauth, et~al.
\newblock Gemini: A family of highly capable multimodal models.
\newblock \emph{arXiv preprint arXiv:2312.11805}, 2023.

\bibitem[Team(2024{\natexlab{b}})]{yardandgarden}
Iowa State~University Team.
\newblock Yard and garden.
\newblock \url{https://yardandgarden.extension.iastate.edu/problems-pests}, 2024{\natexlab{b}}.

\bibitem[Team(2024{\natexlab{c}})]{internvl2}
OpenGVLab Team.
\newblock Internvl2: Better than the best—expanding performance boundaries of open-source multimodal models with the progressive scaling strategy, 2024{\natexlab{c}}.

\bibitem[Team(2024{\natexlab{d}})]{plantvillage}
Plant~Village Team.
\newblock Plant village.
\newblock \url{https://plantvillage.psu.edu/plants}, 2024{\natexlab{d}}.

\bibitem[Team(2024{\natexlab{e}})]{wischorticulture}
Wisconsin~Horticulture Team.
\newblock Wisconsin horticulture.
\newblock \url{https://hort.extension.wisc.edu/ask-a-gardening-question/}, 2024{\natexlab{e}}.

\bibitem[Tong et~al.(2024{\natexlab{a}})Tong, Brown, Wu, Woo, Middepogu, Akula, Yang, Yang, Iyer, Pan, Wang, Fergus, LeCun, and Xie]{cambrian1}
Shengbang Tong, Ellis Brown, Penghao Wu, Sanghyun Woo, Manoj Middepogu, Sai~Charitha Akula, Jihan Yang, Shusheng Yang, Adithya Iyer, Xichen Pan, Austin Wang, Rob Fergus, Yann LeCun, and Saining Xie.
\newblock Cambrian-1: A fully open, vision-centric exploration of multimodal llms.
\newblock \emph{arXiv preprint arXiv:2406.16860}, 2024{\natexlab{a}}.

\bibitem[Tong et~al.(2024{\natexlab{b}})Tong, Liu, Zhai, Ma, LeCun, and Xie]{eyeswideshut}
Shengbang Tong, Zhuang Liu, Yuexiang Zhai, Yi Ma, Yann LeCun, and Saining Xie.
\newblock Eyes wide shut? exploring the visual shortcomings of multimodal {LLMs}.
\newblock In \emph{CVPR}, 2024{\natexlab{b}}.

\bibitem[Tripathi and Maktedar(2020)]{TRIPATHI2020183}
Mukesh~Kumar Tripathi and Dhananjay~D. Maktedar.
\newblock A role of computer vision in fruits and vegetables among various horticulture products of agriculture fields: A survey.
\newblock \emph{Information Processing in Agriculture}, 7\penalty0 (2):\penalty0 183--203, 2020.

\bibitem[Wei et~al.(2024{\natexlab{a}})Wei, Karina, Chung, Jiao, Papay, Glaese, Schulman, and Fedus]{wei2024simpleqa}
Jason Wei, Nguyen Karina, Hyung~Won Chung, Yunxin~Joy Jiao, Spencer Papay, Amelia Glaese, John Schulman, and William Fedus.
\newblock Measuring short-form factuality in large language models.
\newblock \emph{arXiv preprint arXiv:2411.04368}, 2024{\natexlab{a}}.

\bibitem[Wei et~al.(2024{\natexlab{b}})Wei, Chen, Huang, and Yu]{10.1145/3664647.3680599}
Tianqi Wei, Zhi Chen, Zi Huang, and Xin Yu.
\newblock Benchmarking in-the-wild multimodal disease recognition and a versatile baseline.
\newblock In \emph{ACM MM}, 2024{\natexlab{b}}.

\bibitem[xAI Team(2024)]{grok}
xAI Team.
\newblock Grok, 2024.

\bibitem[Yue et~al.(2024)Yue, Ni, Zhang, Zheng, Liu, Zhang, Stevens, Jiang, Ren, Sun, et~al.]{yue2024mmmu}
Xiang Yue, Yuansheng Ni, Kai Zhang, Tianyu Zheng, Ruoqi Liu, Ge Zhang, Samuel Stevens, Dongfu Jiang, Weiming Ren, Yuxuan Sun, et~al.
\newblock Mmmu: A massive multi-discipline multimodal understanding and reasoning benchmark for expert agi.
\newblock In \emph{CVPR}, 2024.

\bibitem[Zhang et~al.(2024{\natexlab{a}})Zhang, Sun, Chen, Liu, Yuan, Zheng, Wang, Yang, Yan, Zhong, et~al.]{zhang2024empowering}
Hang Zhang, Jiawei Sun, Renqi Chen, Wei Liu, Zhonghang Yuan, Xinzhe Zheng, Zhefan Wang, Zhiyuan Yang, Hang Yan, Hansen Zhong, et~al.
\newblock Empowering and assessing the utility of large language models in crop science.
\newblock In \emph{NeurIPS}, 2024{\natexlab{a}}.

\bibitem[Zhang et~al.(2024{\natexlab{b}})Zhang, Ma, Cui, Li, Zhang, and Xie]{ZHANG2024109587}
Kunpeng Zhang, Li Ma, Beibei Cui, Xin Li, Boqiang Zhang, and Na Xie.
\newblock Visual large language model for wheat disease diagnosis in the wild.
\newblock \emph{Computers and Electronics in Agriculture}, 227:\penalty0 109587, 2024{\natexlab{b}}.

\bibitem[Zhang et~al.(2024{\natexlab{c}})Zhang, Chen, Jin, Wang, Ji, Wang, and Han]{zhang2024comprehensive}
Yu Zhang, Xiusi Chen, Bowen Jin, Sheng Wang, Shuiwang Ji, Wei Wang, and Jiawei Han.
\newblock A comprehensive survey of scientific large language models and their applications in scientific discovery.
\newblock \emph{arXiv preprint arXiv:2406.10833}, 2024{\natexlab{c}}.

\bibitem[Zhou and Ryo(2024)]{zhou2024agribench}
Yutong Zhou and Masahiro Ryo.
\newblock Agribench: A hierarchical agriculture benchmark for multimodal large language models.
\newblock \emph{arXiv preprint arXiv:2412.00465}, 2024.

\end{thebibliography}
}

\newpage% \clearpage
\setcounter{page}{1}
% \maketitlesupplementary

\renewcommand{\thesection}{\Alph{section}}
\renewcommand{\thefigure}{\Alph{figure}}
\renewcommand{\thetable}{\Alph{table}}
\setcounter{section}{0}
\setcounter{figure}{0}
\setcounter{table}{0}

% \section*{Supplementary Material}

\etocdepthtag.toc{mtappendix}
\etocsettagdepth{mtchapter}{none}
\etocsettagdepth{mtappendix}{subsection}
{
  \hypersetup{
    linkcolor = black
  }
  \tableofcontents
}

\section{Evaluated Large Vision Language Models}\label{sec:evaluated_vlms}

Our evaluation and analysis are conducted mainly on the group of models listed in Table~2 in the main paper. We have chosen models such that they cover most of the popular and best-performing methods used by recent multimodal understanding work. In this part, we discuss all the models we have used in our experiments and explain their evaluation details, the public checkpoints we have chosen, and display the prompts we used to adapt the model to our datasets. 

During evaluation, we chose to follow the standard prompt provided by the authors whenever possible for multiple-choice and short-answer questions. When the prompt is not provided for the model, we select a custom prompt that is created through several iterations of prompt engineering to select the one that produces the most effective results. The images are always included as the prefix. 

\vspace{1mm}\mypar{Proprietary Models.} We used three proprietary models in our evaluation: GPT-o4-mini~\cite{gpt4v}, Gemini 1.5 Pro~\cite{gemini15}, and Claude 3 Haiku~\cite{claude}. Below we note the model API version used for evaluation. 
\begin{itemize}
    \item GPT-o4-mini: May 13-15, 2025.
    \item Gemini 1.5 Pro: November 1-13, 2024. 
    \item Claude 3 Haiku: November 13-14, 2024. 
\end{itemize}

\vspace{1mm}\mypar{Cambrian-1}~\cite{cambrian1}. Cambrian-1 is a recent state-of-the-art model that excels at visual-centric tasks. This model explores combinations of vision encoders, text and image integration techniques, and instruction tuning strategies. We use the official implementation and checkpoint\footnote{\url{https://github.com/cambrian-mllm/cambrian}} with a LLaMA3-8B-Instruct LLM backbone model in our evaluation. 

\vspace{1mm}\mypar{InternVL2}~\cite{internvl2}. InternVL scales up the vision foundation model while aligning it with the backbone LLM, and is trained on web-scale image-text data to achieve strong performance across a variety of vision-centric tasks. We use the official implementation and checkpoint\footnote{\url{https://huggingface.co/OpenGVLab/InternVL2-8B}} with the InternViT-300M-448px vision backbone and Internlm2.5-7B-chat language backbone in our evaluation. 

\vspace{1mm}\mypar{LLaMA-3.2}~\cite{dubey2024llama3}. LLaMA-3.2 is the first collection of multimodal large language model from the LLaMA family that was previously text-only. The integration of vision involves utilizing cross-attention layers and a pre-trained vision encoder that feeds directly into the text-processor. The model follows a commonly used training recipe that includes pretraining on noisy image-text pairs and then high-quality knowledge enhanced pairs. Notably, the language-model parameters were frozen during the training of alignment of image and text to retain strong text-only capabilities. We use the official implementation and checkpoint\footnote{\url{https://huggingface.co/meta-llama/Llama-3.2-11B-Vision}} that uses a LLaMA-3.1 text-only language backbone in our evaluation. When evaluating the model, we choose to use a custom prompt since no standard prompt is provided. 

\vspace{1mm}\mypar{LLaVA-NeXT}~\cite{llavanext}. LLaVA-NeXT expands on LLaVA by using the same instruction tuning method to give the model the ability to process and reason about multi-images, multi-grames, and multi-views. We use the official implementation and checkpoint\footnote{\url{https://huggingface.co/llava-hf/llama3-llava-next-8b-hf}} with LLaMA-3-8B Instruct as the language backbone in our evaluation. 

\vspace{1mm}\mypar{LLaVA-OneVision}~\cite{llavaonevision}. LLaVA-OneVision builds on LLaVA-NeXT with the capability to analyze single images, multi-images, and video scenarios. Most impressively, it allows for video understanding through task transfer from images but this is not explored in our evaluation. We use the official implementation and checkpoint\footnote{\url{https://huggingface.co/llava-hf/llava-onevision-qwen2-7b-ov-hf}} that uses a base architecture consisting of SigLIP-SO400M-Patch14-384 and Qwen2-7b in our evaluation. 

\vspace{1mm}\mypar{LLaVA-1.5-7B / LLaVA-1.5-13B}~\cite{llava}. LLaVA introduces the idea of instruction tuning a multimodal model with GPT-4 generated instruction-following data for associated images. This gives it the ability to achieve impressive abilities to act as an instruction-following general agent. We use the official implementation and checkpoints\footnote{\url{https://huggingface.co/llava-hf/llava-1.5-7b-hf}}\footnote{\url{https://huggingface.co/llava-hf/llava-1.5-13b-hf}} with a CLIP ViT-L/14 vision backbone and Vicuna1.5-7B / Vicuna1.5-13B in our evaluation. 

\vspace{1mm}\mypar{Qwen-VL-7B}~\cite{Qwen-VL}. Qwen-VL is a large vision language model that has the ability to perform various vision-language tasks including image captioning, visual grounding and more, not only limited to question answering. This model is multi-lingual in Chinese and English and was pre-trained using an interleaved image-text technique. We use the official implementation and checkpoint\footnote{\url{https://huggingface.co/Qwen/Qwen-VL}} that uses Qwen-7B as the language backbone and CLIP ViT bigG/14 as the vision encoder in our evaluation. 

\vspace{1mm}\mypar{VILA1.5-13B}~\cite{vila}. VILA is trained using an enhanced pre-training method that involves interleaved visual language data. Additionally, during the supervised fine-tuning stage, the data includes text-only instruction data to help the model retain strong text-only capabilities. We use the official implementation and checkpoint\footnote{\url{https://github.com/NVlabs/VILA}} with a LLaMA3-8B LLM backbone and SigLIP-SO400M-Patch14-384 vision encoder in our evaluation.

\section{Dataset Curation Details} \label{sec:dataset_curation_details}

This section outlines the multi-stage curation pipeline of \ourwork and describes the prompts designed for each question type and subdomain. 

\begin{figure*}
    \centering
    \includegraphics[width=0.99\linewidth]{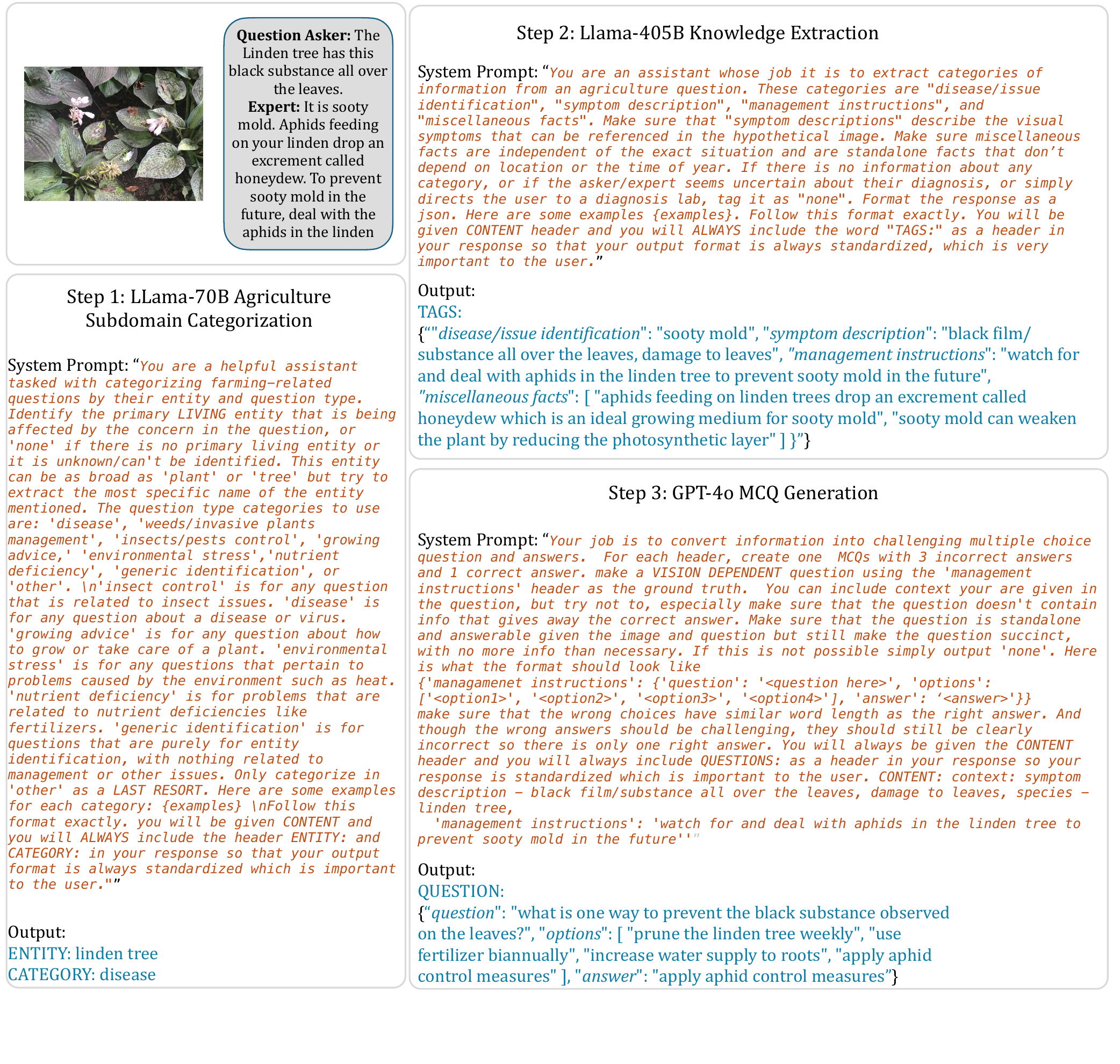}
    \vspace{-10mm}
    \captionof{figure}{Prompts used in different stages of our data curation pipeline.}
    \label{fig:supp:workflow}
    \vspace{-4mm}
\end{figure*}

\begin{figure*}
    \centering
    \includegraphics[width=0.91\linewidth]{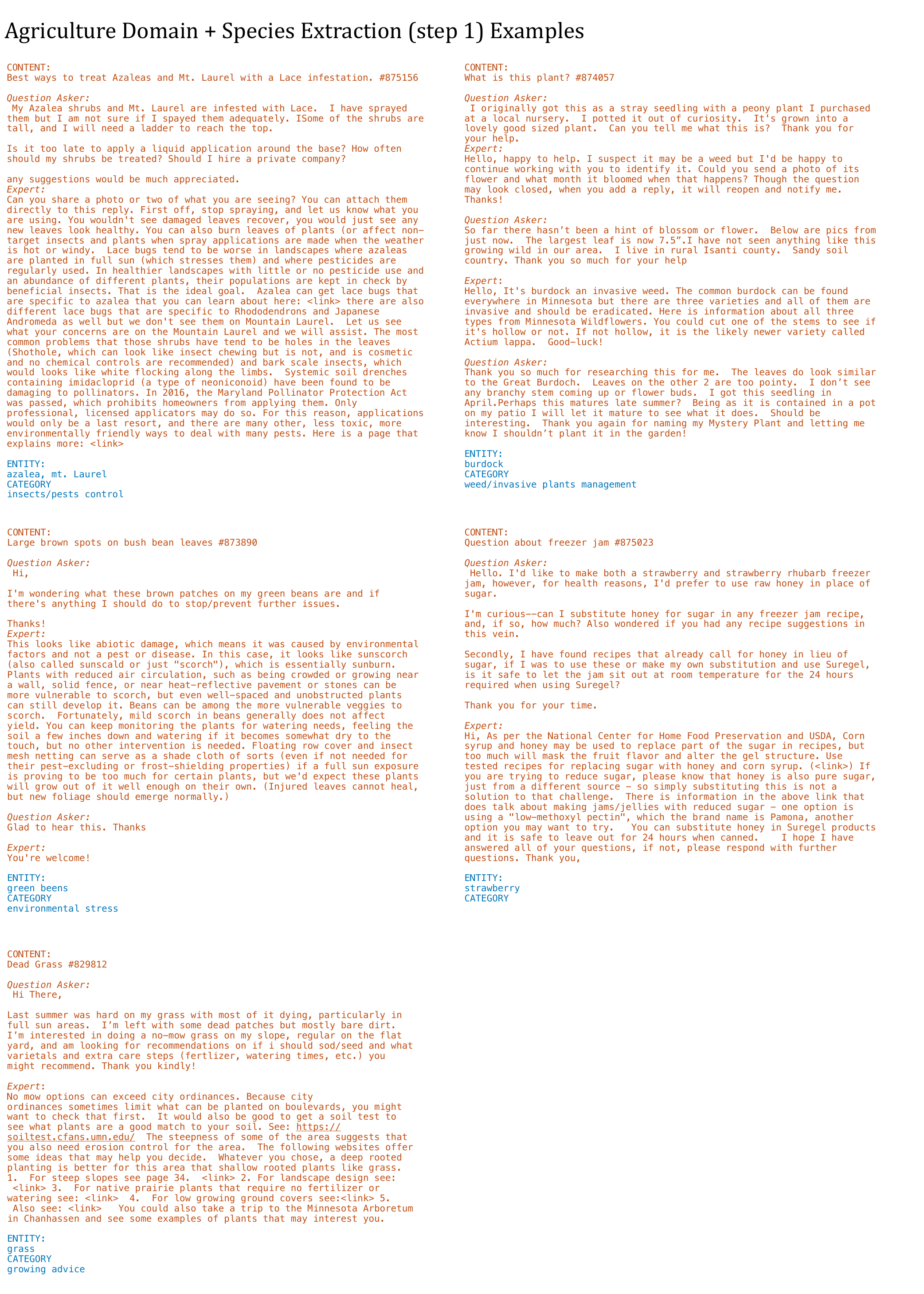}
    \vspace{-4mm}
    \captionof{figure}{Examples included in prompt during the agriculture domain categorization (step 1).}
    \label{fig:supp:qtype_examples}
    \vspace{-4mm}
\end{figure*}

\begin{figure*}
    \centering
    \includegraphics[width=0.93\linewidth]{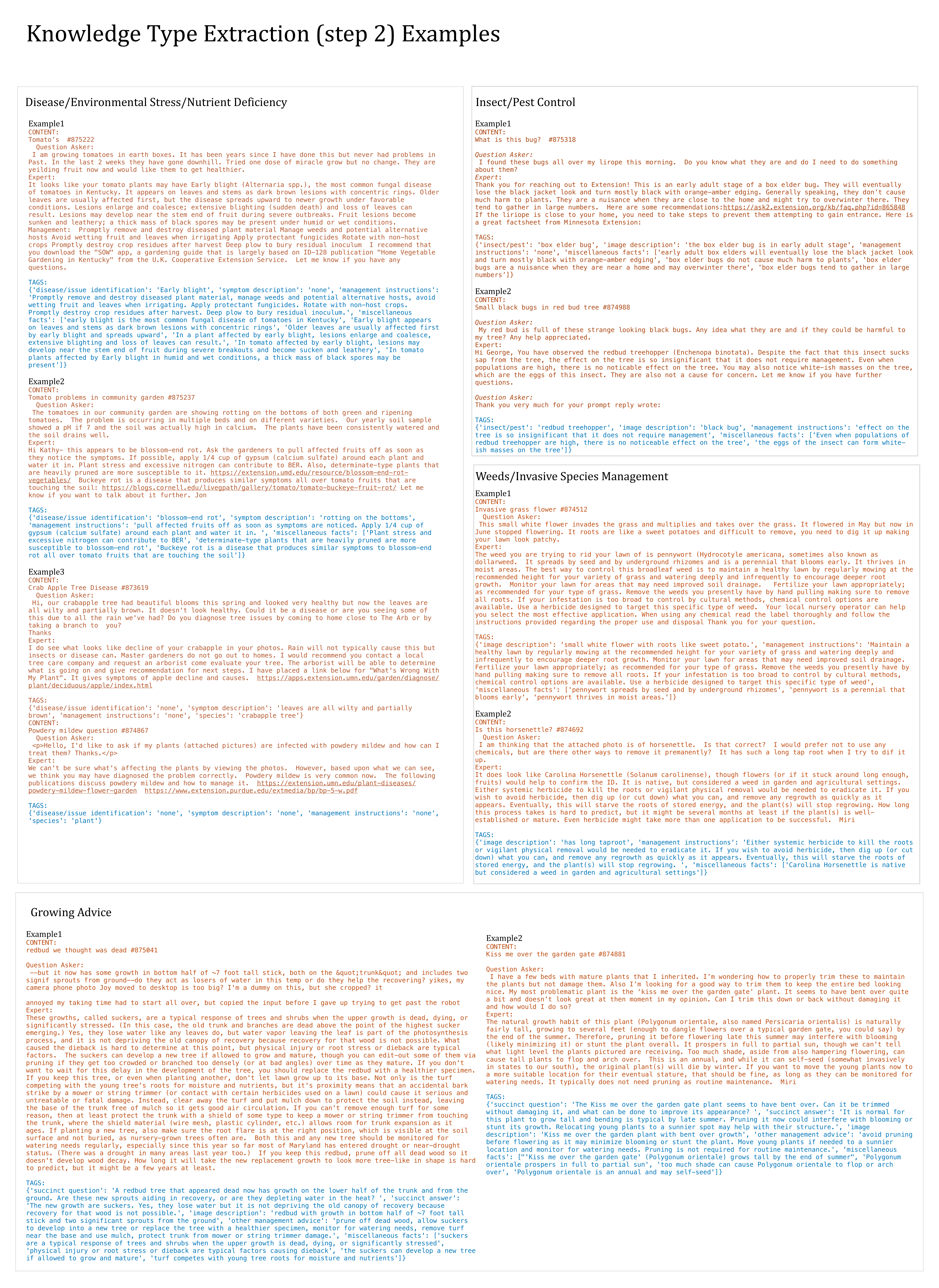}
    \vspace{-4mm}
    \captionof{figure}{Examples included in prompt during the knowledge extraction (step 2) based on agriculture subdomain type.}
    \label{fig:supp:info_extract_examples}
    \vspace{-4mm}
\end{figure*}

\begin{figure*}[!t]
    \centering
    \includegraphics[trim=0 60 0 10, clip, width=0.99\linewidth]{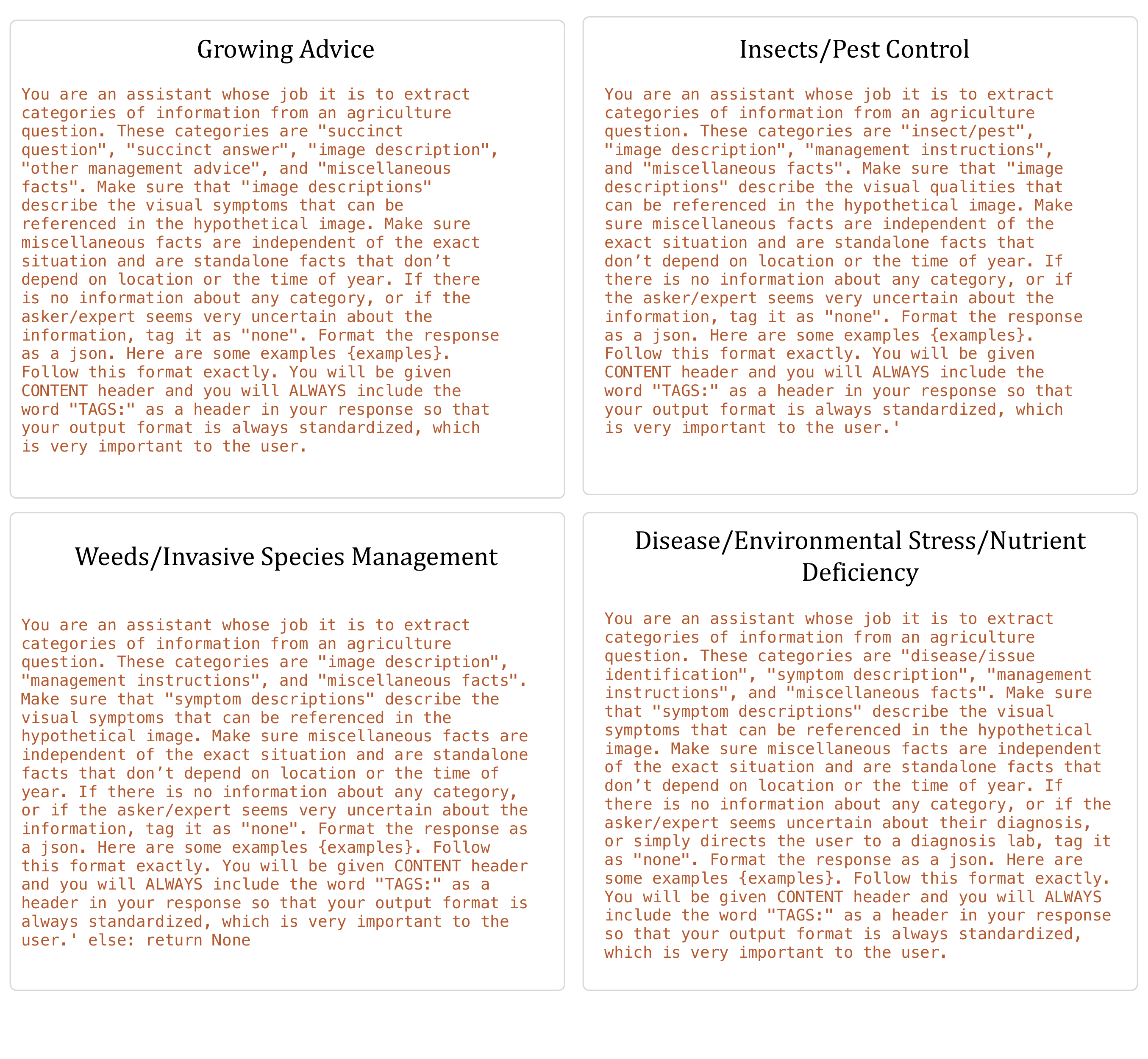}
    \vspace{-3mm}
    \captionof{figure}{Prompts used for each subdomain during knowledge extraction (step 2).}
    \label{fig:supp:info_extraction_prompts}
    \vspace{-4mm}
\end{figure*}

% \begin{figure*}
%     \centering
%     \includegraphics[width=0.99\linewidth]{figures/pdf/supplementals/short_oeq_prompts.pdf}
%     % \vspace{-10mm}
%     \captionof{figure}{Prompts for our LLM-as-judge on short-answer OEQs.}
%     \label{fig:supp:short_oeq}
%     \vspace{-4mm}
% \end{figure*}

\begin{figure*}
    \centering
    \includegraphics[width=0.99\linewidth]{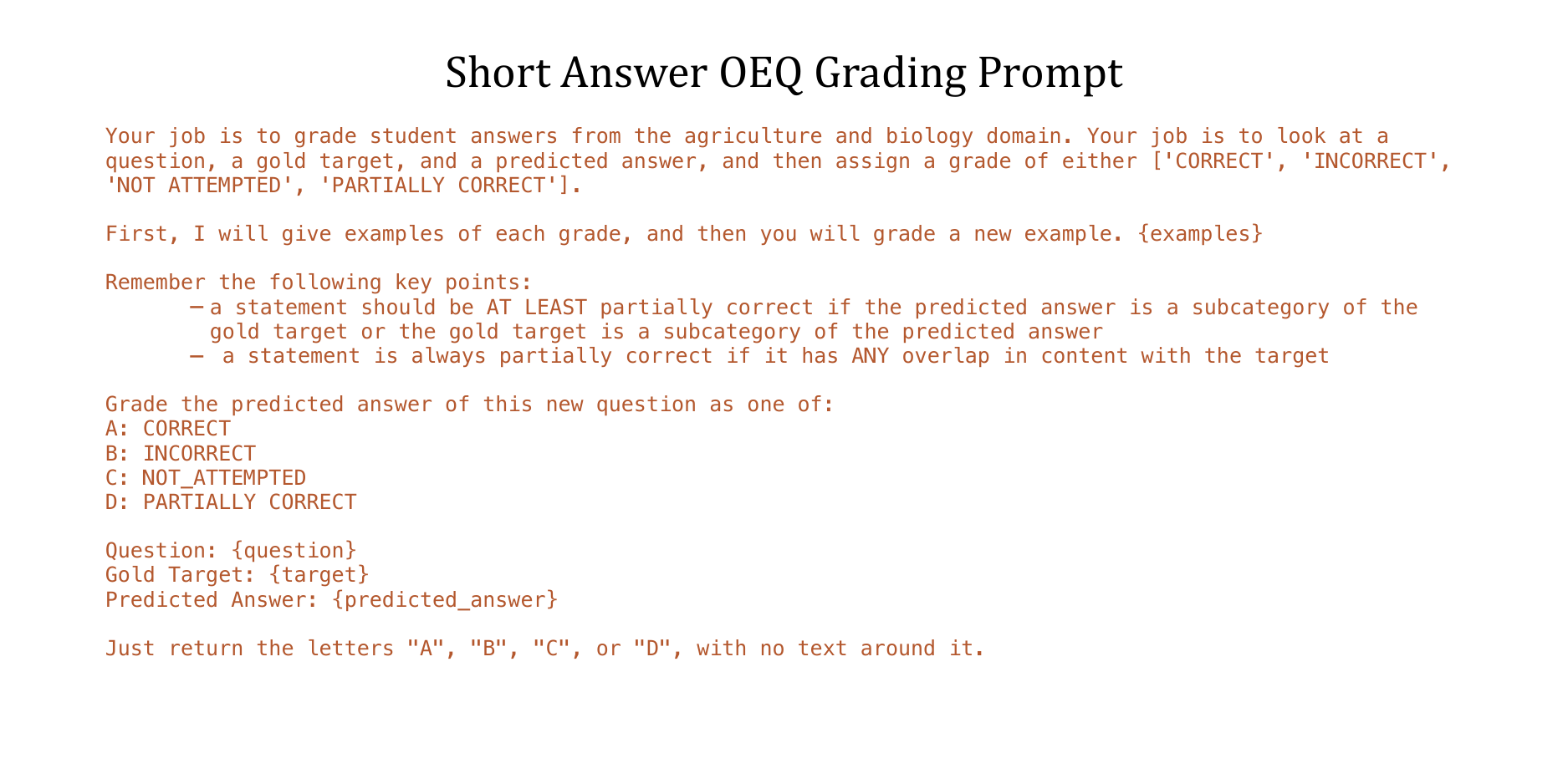}
    % \vspace{-13mm}
    \captionof{figure}{Grading prompt for our LLM-as-judge on short-answer OEQ.}
    \label{fig:supp:short_oeq1}
    \vspace{-4mm}
\end{figure*}

\begin{figure*}
    \centering
    \includegraphics[width=0.99\linewidth]{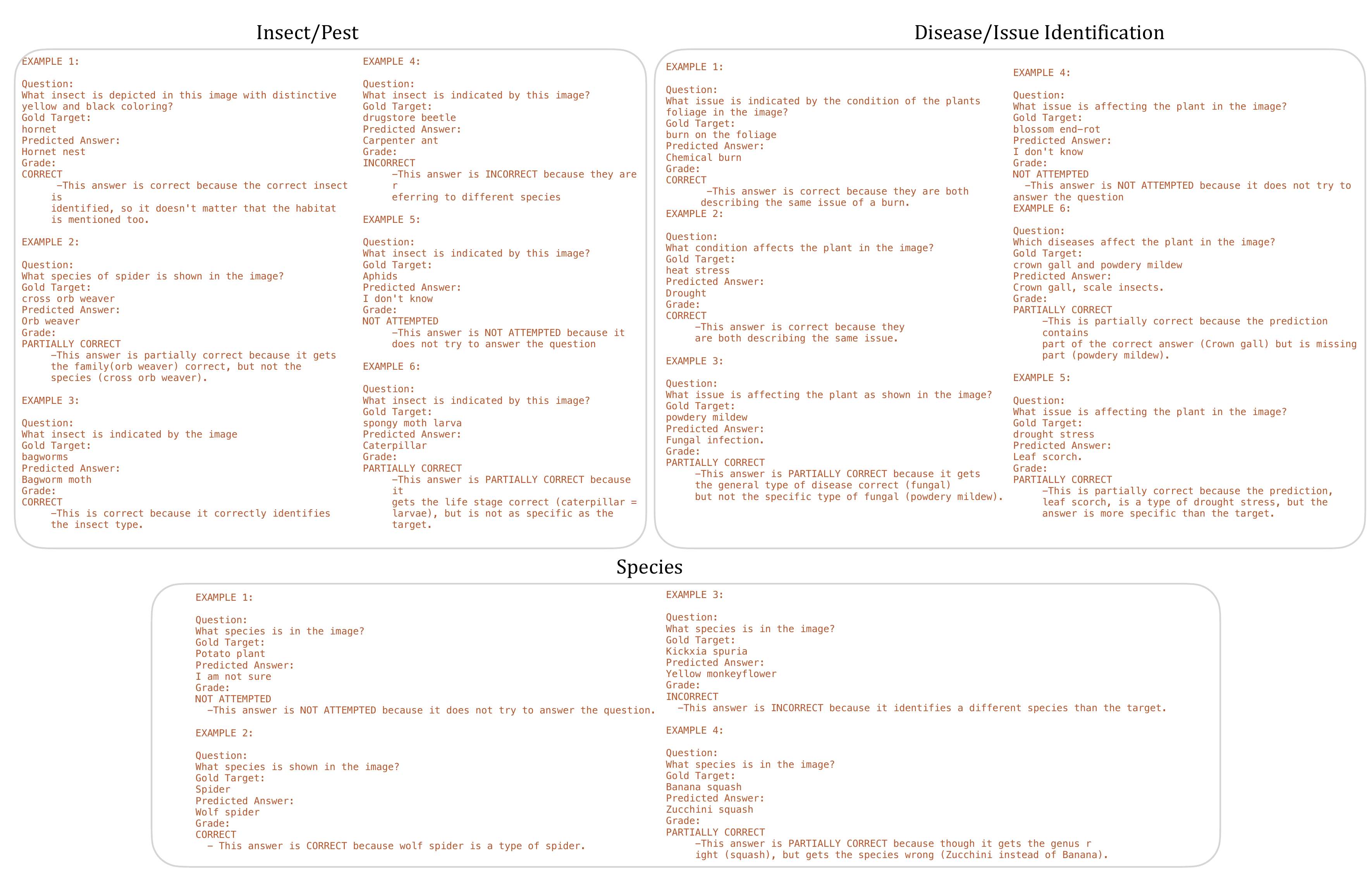}
    % \vspace{-13mm}
    \captionof{figure}{Unique examples included for short-answer categories added to the grading prompt for our LLM-as-judge on short answer OEQ.}
    \label{fig:supp:short_oeq2}
    \vspace{-4mm}
\end{figure*}
\begin{figure*}
    \centering
    \includegraphics[width=0.99\linewidth]{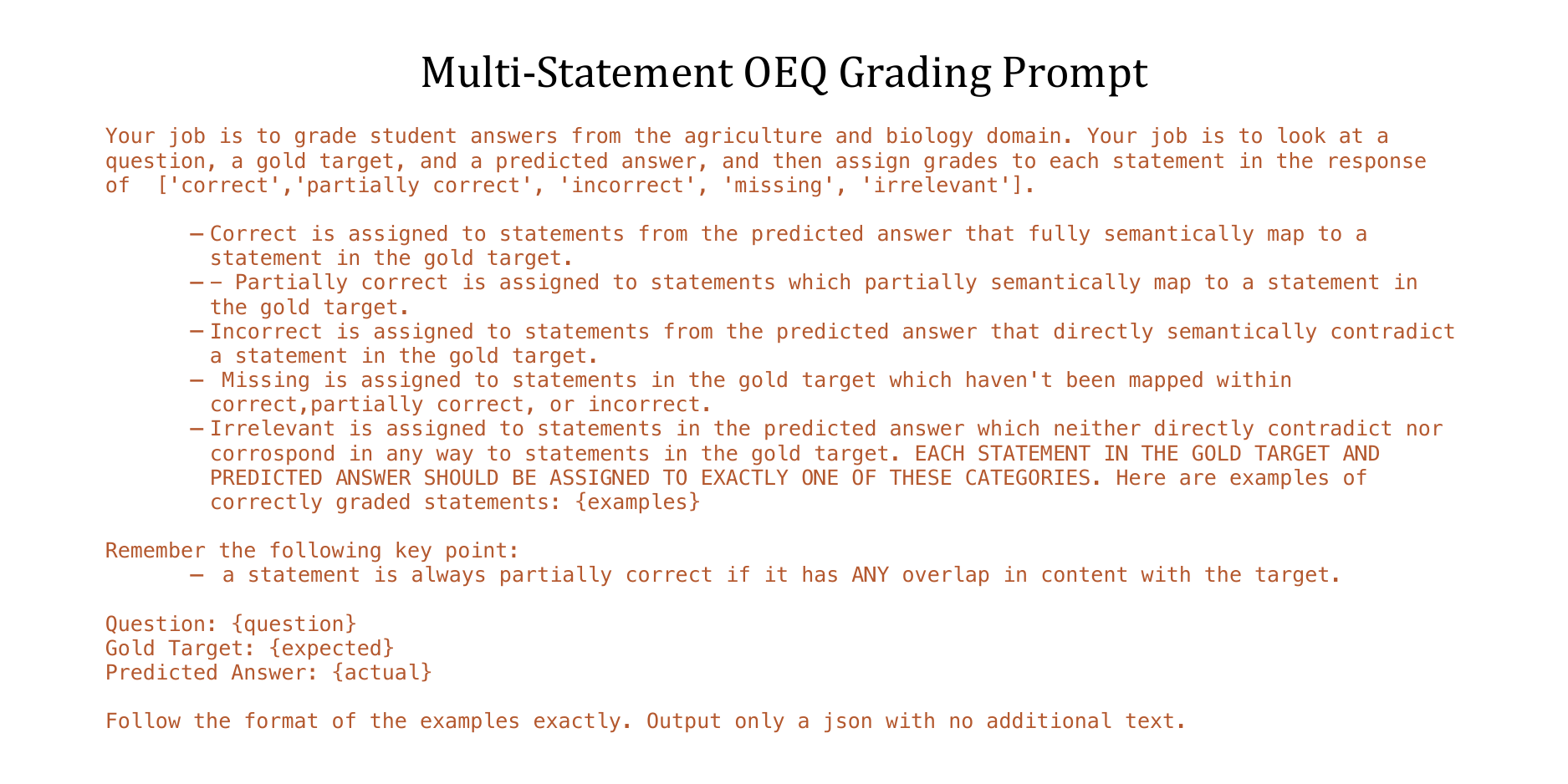}
    % \vspace{-13mm}
    \captionof{figure}{Prompt for categorizing statements in our LLM-as-judge on multi-sentence (long-answer) OEQ.}
    \label{fig:supp:long_oeq1}
    \vspace{-4mm}
\end{figure*}

\begin{figure*}
    \centering
    \includegraphics[width=0.99\linewidth]{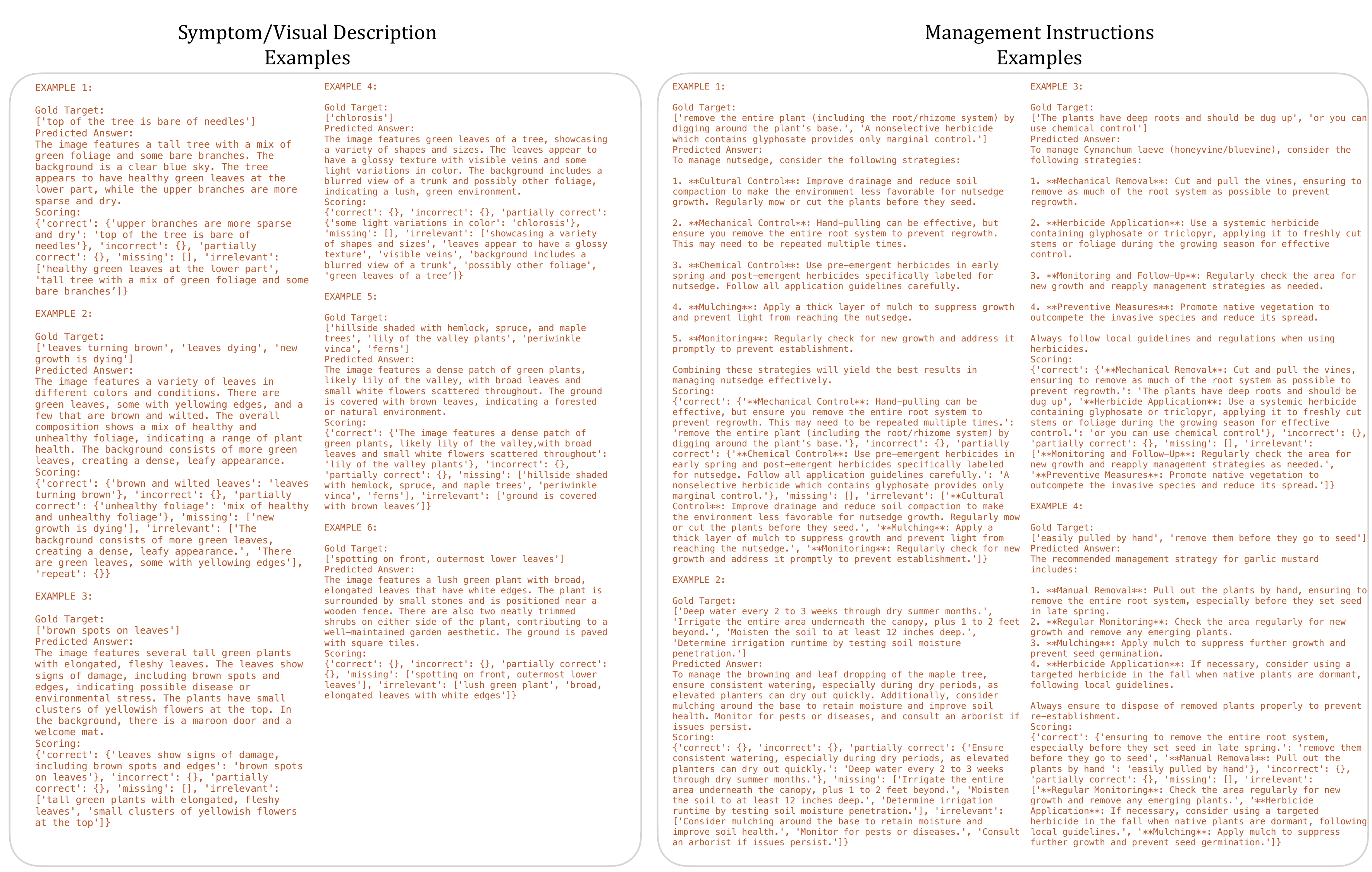}
    % \vspace{-13mm}
    \captionof{figure}{Unique examples included for multi-statement categories added to the grading prompt for our LLM-as-judge on multi-sentence (long-answer) OEQ.}
    \label{fig:supp:long_oeq2}
    \vspace{-4mm}
\end{figure*}

\subsection{Stage 1: Question Categorization}

In the first step, we employ the Llama-70B model~\cite{dubey2024llama3} to categorize questions into predefined agriculture subdomains while identifying the primary living entity affected by the query. Our systematically crafted prompt (Figure~\ref{fig:supp:workflow}) guides the model to extract the most specific living entity mentioned, such as ``apple tree'' or ``honeybee,'' or to assign ``none'' when the entity is unclear or absent.

The subdomains include \textit{Disease, Weeds/Invasive Plant Management, Insect/Pest Control, Growing Advice, Environmental Stress, Nutrient Deficiency, Generic Identification}, and \textit{Other}. Each subdomain is succinctly defined within the prompt, with illustrative examples provided in Figure~\ref{fig:supp:qtype_examples} to address ambiguous or edge-case scenarios. The prompt enforces a standardized output format, ensuring consistency with the inclusion of ``ENTITY:'' and ``CATEGORY:'' headers.

To enhance robustness, the prompt includes examples of complex or overlapping cases, ensuring accurate classification even for questions that span multiple subdomains or lack explicit details. By embedding these clarifications, the design supports reliable categorization across diverse agricultural contexts. 

\subsection{Stage 2: Information Extraction}

In the second step, we design prompts to extract granular categories of information from agricultural questions. These categories are tailored to the specific subdomain identified in Step 1, ensuring that the extracted information is both relevant and actionable. 

\vspace{1mm}\mypar{Weeds/Invasive Plants Management.} For the ``weeds/invasive plants management'' subdomain, the extraction focuses on: (1) \textit{Image Description}, visual characteristics of the weed or invasive plant, (2) \textit{Management Instructions}, actionable strategies for control, and (3) \textit{Miscellaneous Facts}, contextual expert insights. The name of the weed itself is already extracted in Step 1. This ensures that the emphasis remains on descriptions, actionable measures, and expert knowledge.

\vspace{1mm}\mypar{Insects/Pests Control.} For this subdomain, the categories include: (1) \textit{Insect/Pest}, identifying the pest in focus, (2) \textit{Image Description}, visual traits of the pest or evidence of damage, (3) \textit{Management Instructions}, guidance for mitigation, and (4) \textit{Miscellaneous Facts}, contextual expert insights. The primary plant affected, if exists, is identified in Step 1, thus this step concentrates on pest-specific details, such as visual features or damage patterns, and the corresponding management strategies.

\vspace{1mm}\mypar{Nutrient Deficiency, Disease, Environmental Stress.} For these subdomains, we group them due to shared characteristics. The extracted categories are: (1) \textit{Disease/Issue Identification}, specifying the underlying cause, (2) \textit{Symptom Description}, observable signs such as discoloration or stunted growth, (3) \textit{Management Instructions}, remediation or prevention strategies, and (4) \textit{Miscellaneous Facts}, contextual expert insights. These subdomains are defined by their symptomatic presentation, the underlying conditions, and the need for targeted management interventions.

\vspace{1mm}\mypar{Growing Advice.} For this subdomain, the variability in question structure necessitates tailored extractions: (1) \textit{Succinct Question}, a concise reformulation of the user query, (2) \textit{Succinct Answer}, a precise response to the query, (3) \textit{Image Description}, any relevant visual details, and (4) \textit{Miscellaneous Facts}, contextual expert insights.

Importantly, besides distinguishing the extraction types, we also put different examples of pre-made knowledge extraction into the prompt, see Figure~\ref{fig:supp:info_extract_examples}. Prompts given to the model for each subdomain can be seen in Figure~\ref{fig:supp:info_extraction_prompts}. 

The \textit{Miscellaneous Facts} category is extracted across all subdomains but is not directly used in subsequent steps. Instead, it captures standalone expert information that can contextualize a user’s issue.

To optimize extraction accuracy, we distinguish between ``Symptom Description'' (used for nutrient deficiency, disease, and environmental stress) and ``Image Description'' (used for weeds/invasive plants and insects/pests). While these serve a similar purpose—capturing observable or visual details—they are unified under the term ``Symptom/Visual Description'' in subsequent steps to maintain consistency.

\subsection{Stage 3: Question Generation} 

In the final step, the extracted agricultural facts are transformed into evaluative question-answer (QA) pairs, comprising multiple-choice questions (MCQs) and open-ended questions (OEQs) generated using GPT-4o. To enhance relevance, we exclude two knowledge types: (1) \textit{Growing Advice}, as image content often lacks direct correlation with the user's issue, and (2) \textit{Miscellaneous Facts}, since these provide general context but do not directly relate to the user's image. This refinement narrows the scope to five key knowledge types for downstream processing, including \textit{Disease/Issue Identification}, \textit{Symptom/Visual Description}, \textit{Management Instructions}, \textit{Insect/Pest}, and \textit{Species}. 

To ensure clarity and relevance, we employ a standardized prompt structure (see Figure~\ref{fig:supp:workflow} tailored to each knowledge type. While the core structure remains consistent, the phrasing explicitly references the specific knowledge type being addressed. This targeted design allows the prompts to focus on generating well-contextualized and relevant questions. For added precision, the prompts incorporate contextual details where applicable: (1) For \textit{species-related questions}, only symptom/visual description information is referenced, ensuring the focus remains on observable traits, and (2) for \textit{symptom/visual-related questions}, species information is used to provide context, helping to ground the questions in specific agricultural scenarios.

This contextualization ensures that the generated questions integrate both user-provided information and extracted context seamlessly. The result is a set of comprehensive and ``fair'' evaluative questions, designed to effectively assess multimodal agricultural understanding.

\begin{figure*}[!t]
    \centering
    \includegraphics[trim=0 0 0 80, clip, width=0.8\linewidth]{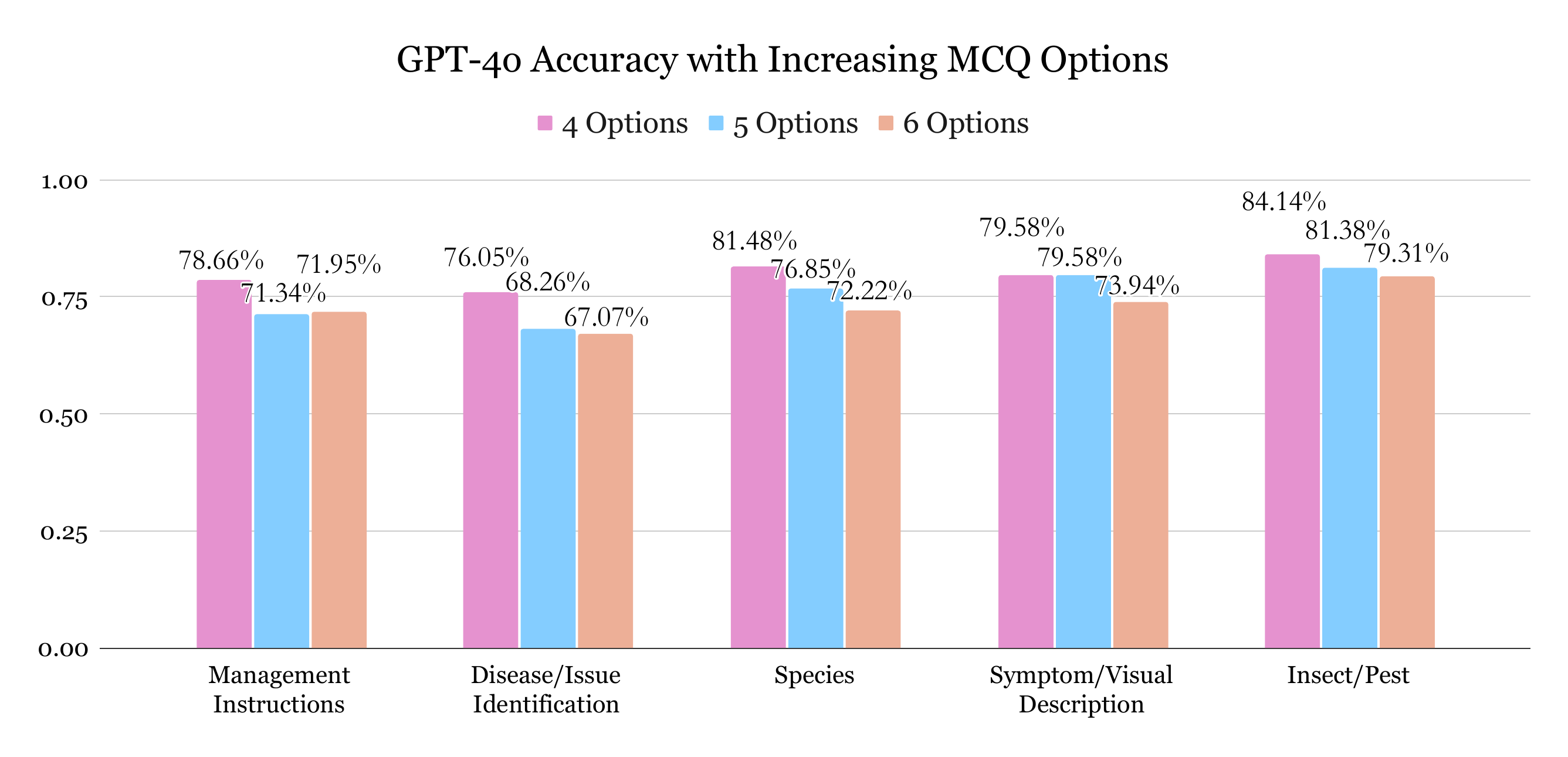}
    \vspace{-4mm}
    \captionof{figure}{\textbf{GPT-4o accuracy with increasing MCQ options.} Model performance on MCQs across different categories, comparing accuracy scores when varying the number of answer options (4, 5, and 6). We observe a 5-10\% difference in accuracy across categories between the 4-option and 6-option configurations, with performance generally decreasing as the number of options increases. }
    \label{fig:experiment1}
    \vspace{-4mm}
\end{figure*}

\subsection{Final Stage: Human Verification} 

To guarantee the quality of the evaluation questions, we implemented a human verification process that validates faithfulness, certainty, quality, and MCQ feasibility. The data was distributed through an HTML file containing \ourwork questions and answers, original user questions, expert answers, and corresponding images. Each annotator was given a corresponding Excel file where the user just has to mark false (uncheck the box) for each condition not met per question. To further assist the annotator, we provided a few complex examples of questions that meet and do not meet the requirements, functioning as in-context examples. After collecting these data, only the completely unproblematic ones (all boxes remain checked) were kept. 

\textbf{Faithfulness: } \textit{Do you think the question, ground-truth, and context extract faithful information from the original farmer question?} Our questions are directly based on the original questions and this step functions as a sanity check ensuring the quality of our dataset. The annotator needs to read through the question and the original conversations between the user and the expert.
%We also debated on passing out merch, but quickly realized it is ,ore difficult than expexted. 

\textbf{Certainty: }\textit{Is the expert certain about the answer?} As our ground truth answers are extracted from the expert answers, we only want to include those that are very certain. A higher certainty from the expert means that it is more likely to be correct. We observe that the behaviors of the annotator are to read the responses from the expert and look for keywords like ``may,'' ``not sure,'' ``you have to go to a lab for further inspection.''

\textbf{Quality: }\textit{Are the images suitable for answering the questions?} (Images are not in low-resolution, blurs, pure blackness, etc.) \textit{Are the image/symptom descriptions visible in the presented images?} As our benchmark attempts to evaluate the visual understanding of the models, our human verification removes the questions that do not depend on the images and those with broken images.  For example, it is not fair for our question to ask about the fruit of the plant when the submitted photos only capture the leaves of the tree or the image is blurry. 

\textbf{Feasibility: }\textit{Are all of the wrong choices wrong?} The incorrect choices were generated with GPT-4o, so we need to check to ensure there are no multiple correct answers or an answer that overlaps with the correct answer and remove. For example, the wrong choice might be the common name of a species displayed by the scientific name in the ground truth.

\section{More Evaluation, Implementation, and Design Choices} \label{sec:more_experiments}

\begin{figure*}
    \centering
    \includegraphics[width=0.99\linewidth]{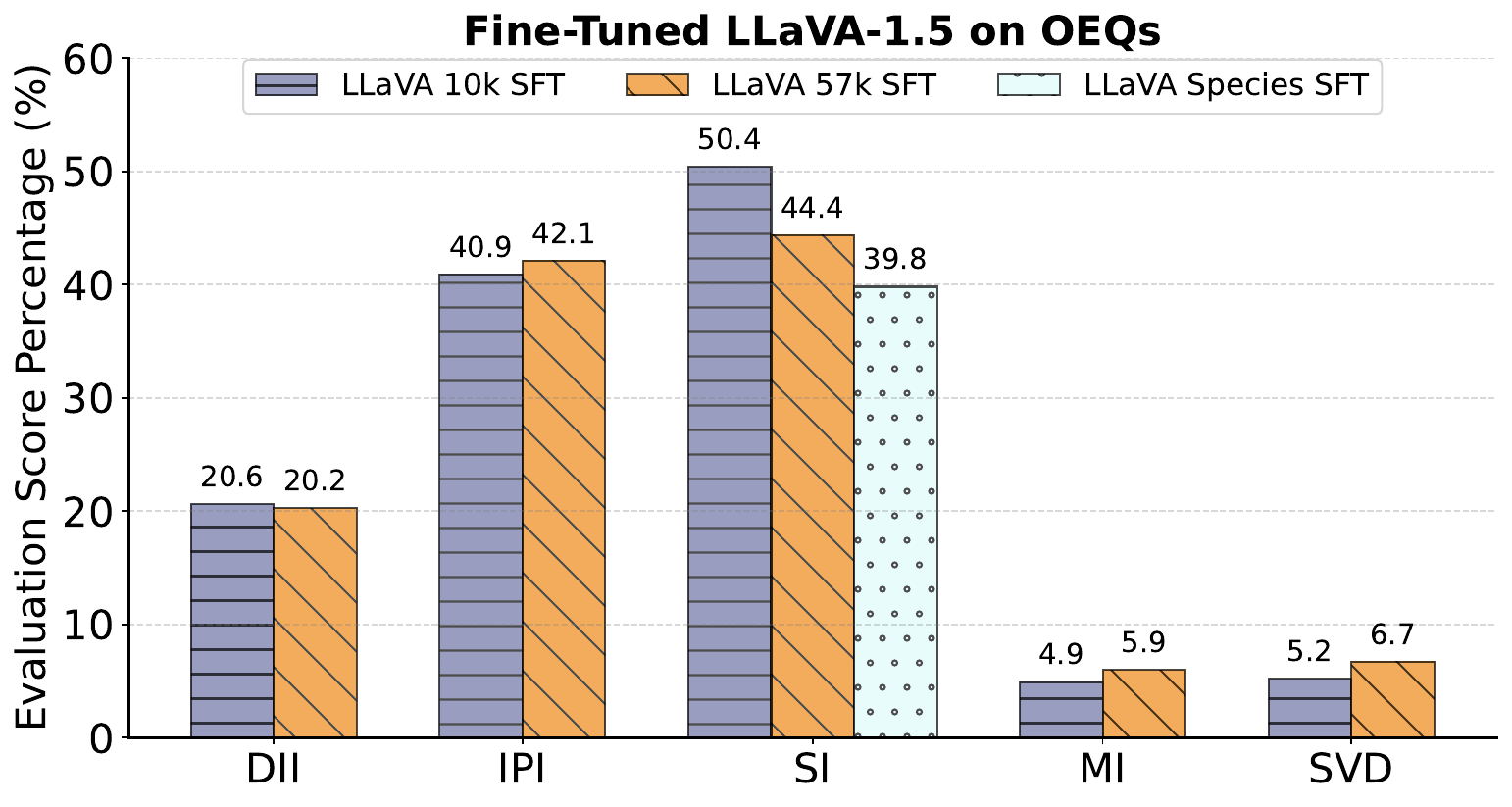}
    % \vspace{-13mm}
    \captionof{figure}{Evaluation scores across five error categories for three fine-tuning setups using the \textsc{AgBase} dataset. Models are fine-tuned on: (1) \textbf{LLaVA 10k SFT}—a mix of \textsc{AgBase} and 10k LLaVA-Instruct samples, (2) \textbf{LLaVA 57k SFT}—a 50-50 blend of \textsc{AgBase} and LLaVA’s original 57k SFT samples, and (3) \textbf{LLaVA Species SFT}—a specialized set focused on species identification with contextual augmentation.}
    \label{fig:supp:finetuning}
    \vspace{-4mm}
\end{figure*}

\mypar{LLM-as-judge.} To perform evaluation on few-word and multi-statement OEQ responses, we implemented the LLM-as-judge methodology using GPT-4.1. Our prompts for few-word responses (Figure~\ref{fig:supp:short_oeq1},~\ref{fig:supp:short_oeq2}) and multi-statement responses (Figure~\ref{fig:supp:long_oeq1}, \ref{fig:supp:long_oeq2}) contain several in-context examples based on the question category to guide the LLM to correctly categorize the answer as “correct,” “incorrect,” “partially correct,” and “irrelevant.” 

\vspace{1mm}
\mypar{Number of MCQ options.} To determine the optimal number of answer choices for our MCQs, we conducted an ablation study comparing GPT-4o's accuracy when presented with four, five, and six options. For efficiency, we conducted this experiment on a subset of 821 questions, generating 5 wrong answers with GPT-4o. We randomly choose 3, and 4 wrong answers, for the four-choice and five-choice experiment, respectively, and take all choices for the six-choice experiment. While this limited subset may not capture the full variability of the dataset, it provides sufficient evidence to inform our design decisions. Due to the risk of process of elimination with MCQs, we believe that OEQs more accurately capture model performance.

The results, shown in Figure ~\ref{fig:experiment1}, indicate that accuracy decreases as the number of answer choices increases. Specifically, we observed a 5-10\% reduction in accuracy between the 4-option and 6-option configurations. This trend suggests that the model might rely on a process of elimination when selecting an answer, making it more challenging to identify the correct response as the number of options increases. While the decrease in accuracy is not overly significant, we think it justifies our choice to use four options for MCQs.

\vspace{1mm}\mypar{Implementation of \textsc{AgBase} fine-tuning.} 
We fine-tune the LLaVA-v1.5-7B model using a LoRA-based setup. The training is performed with a learning rate of 2e-4, without weight decay, and a cosine learning rate schedule with a 3\% warm-up ratio. We use a per-device batch size of 16 with gradient accumulation steps set to 2, resulting in an effective batch size of 64. The model is trained over 2 epochs using 2 NVIDIA A6000 GPUs.

Dataset preparation involved curating structured multi-turn conversations from a horticultural FAQ knowledge base, paired with user-uploaded images. From an initial pool of 367,331 QA-image pairs, we filtered out questions that had a species value in [\textit{tree}, \textit{bee}, \textit{shrub}, \textit{weed}, \textit{wasp}, \textit{plant}, \textit{insect}, \textit{grass}, \textit{none}, \textit{moth}, \textit{beetle}, \textit{snake}, \textit{caterpillar}, \textit{spider}, \textit{ant}, \textit{mushroom}, \textit{fungus}], because we observe that questions with these common non-species species extractions often contain vague or uncertain examples. This gives us a high-quality dataset of 57,079 samples. Considering the influence of data mixture for training VLMs, we conduct three fine-tuning experiments. (1) The first experiment involves fine-tuning on a combination of our domain-specific dataset, \textsc{AgBase}, and 10,000 samples from LLaVA’s original instruction-tuning dataset, LLaVA-Instruct-150K. (2) The second experiment employs a 50-50 mixture of \textsc{AgBase} by using 57,079 samples from LLaVA’s original SFT set~\cite{llava} (3) The third experiment focuses solely on species identification and consists of 18,109 QA pairs constructed by prepending the full original user queries to 33,777 generic identification samples, allowing us to test the effect of user context on classification accuracy.

In Figure~\ref{fig:supp:finetuning}, we find that the LLaVA 10k SFT model achieves a slightly higher overall accuracy (0.25) compared to the LLaVA 57k SFT model (0.24), suggesting that a smaller, well-curated dataset mixed with domain-specific data may be more effective than a larger, more generic one for knowledge-intensive domain fine-tuning. Additionally, the LLaVA Species SFT model, which includes added user query context for species identification, performs worse than the other models in the species category
, indicating that this additional context provides limited benefit for classification accuracy.
% In all experiments, the questions were hard-coded based on our extracted knowledge categories (detailed in cite here). 

\section{More Dataset Visualization} \label{sec:more_visualization}

\begin{figure*}[!t]
    \centering
    \includegraphics[width=0.99\linewidth]{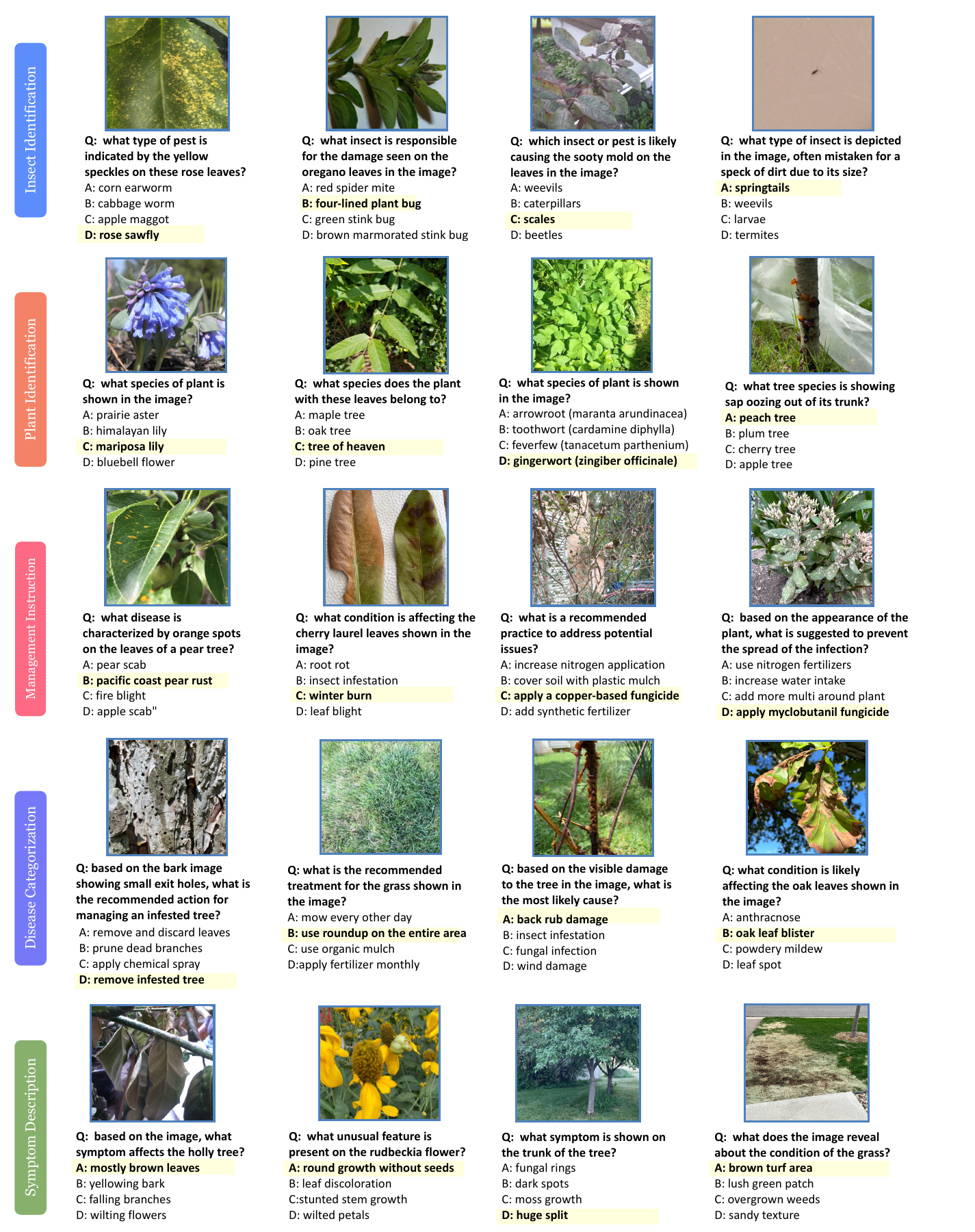}
    \vspace{-3mm}
    \captionof{figure}{Additional visualization of samples in \ourwork. Ground truth selections of each question are highlighted in yellow.}
    \label{fig:supp:supplementary_teaser}
    \vspace{-4mm}
\end{figure*}

In Figure~\ref{fig:supp:supplementary_teaser}, we demonstrate more samples in \ourwork with questions and \textbf{multiple choice answers}.

\begin{figure*}[!t]
    \centering
    \includegraphics[width=0.99\linewidth]{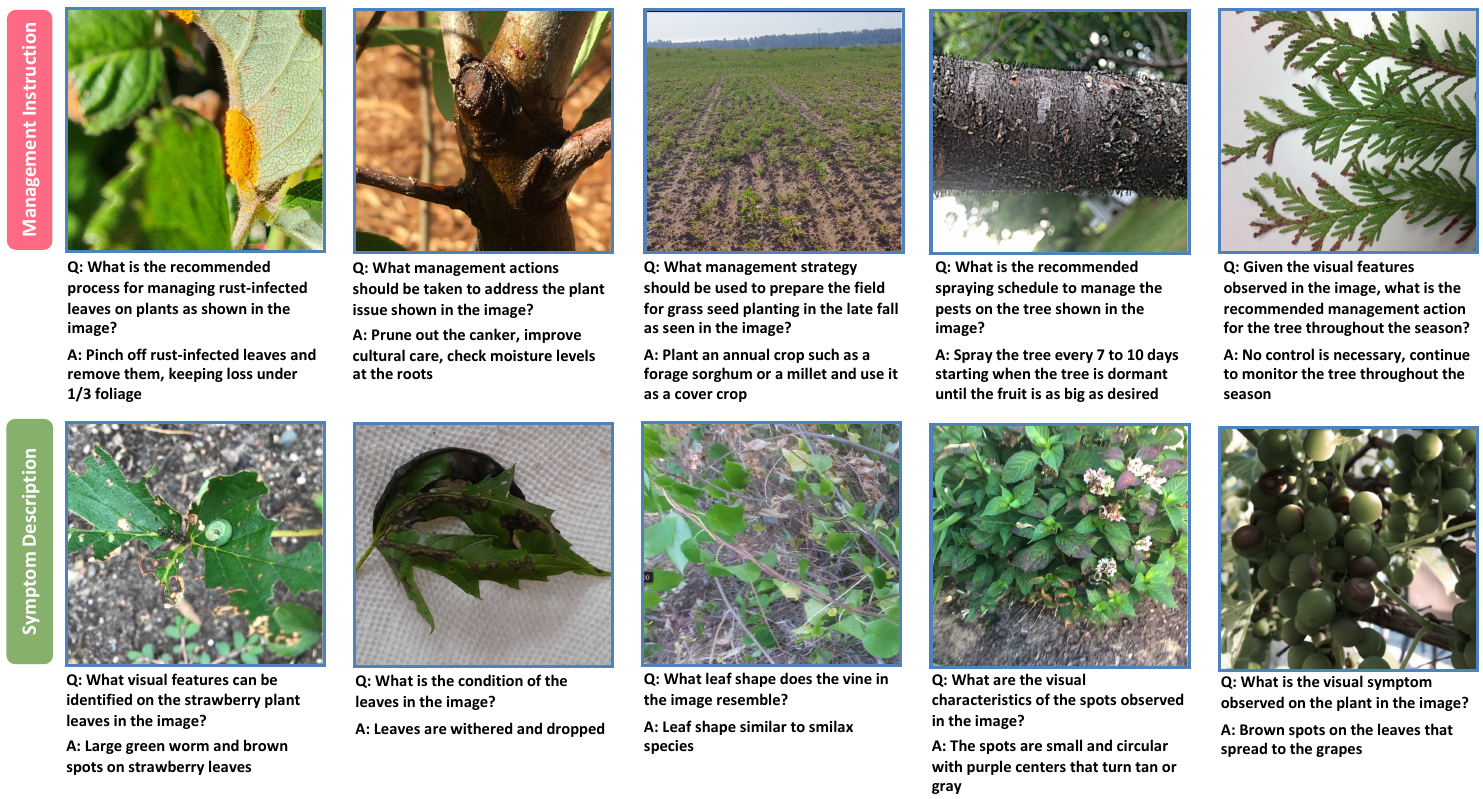}
    \vspace{-3mm}
    \captionof{figure}{Additional visualization of OEQ samples in \ourwork. }
    \label{fig:supp:supplementary_teaser_oeq}
    \vspace{-4mm}
\end{figure*}

In Figure~\ref{fig:supp:supplementary_teaser_oeq}, we demonstrate more samples in \ourwork with \textbf{open-ended questions} and responses. We especially emphasize the long-form responses required from the model for symptom description and management instructions, normally containing multiple facts.

\section{Limitations and Future Work} \label{sec:limitations}

While our work makes unique contributions to agricultural benchmark development and VLM evaluation through knowledge-intensive tasks, we acknowledge several limitations and identify promising directions for future research in this section.

\vspace{1mm}\mypar{Advanced Utilization of Training Data.} Although our curated dataset, \textsc{AgBase}, has proven significant effectiveness for fine-tuning VLMs~\cite{llava} as shown in Section~4 and Figure~6, its potential extends beyond our current usage. As a comprehensive knowledge repository, the dataset presents opportunities for knowledge retrieval and augmented generation (RAG) approaches~\cite{balaguer2024rag}. In particular, the development of vision-centric multimodal RAG systems remains an under-explored yet promising direction. This alternative could enable more effective knowledge extraction and utilization from our dataset, potentially improving model performance on agricultural understanding tasks. We leave the exploration of these advanced techniques for future work.

\vspace{1mm}\mypar{Expanded Model Coverage and Evaluation Protocols.} While our current study encompasses several state-of-the-art and most commonly used VLMs for zero-shot evaluation and fine-tuning analysis, we acknowledge that they represent only a subset of available multimodal architectures and methodologies. To enhance the robustness and generalizability of our findings, we plan to incorporate a broader spectrum of VLMs. Additionally, we plan to conduct more extensive ablation studies and comparative analyses across different model scales and architectures. This comprehensive evaluation will provide deeper insights into the relative strengths and limitations of various approaches in agricultural understanding tasks.

\section{Societal Impact} \label{sec:societal}

\noindent We anticipate no direct negative societal impact of our work. Our dataset is ethically designed, respecting the privacy of Extension.org users by removing personal identifying information such as name, gender, username, and location. Additionally, we have verified to the best of our ability to ensure the removal of images that contain human faces. During dataset curation, we put in great effort to eliminate bias by creating a dataset representative of the original Extension.org questions as well as a balanced dataset across all question types. 

\vspace{1mm}
\mypar{Positive Impact:} We hope that the creation and release of this challenging vision-knowledge intensive dataset can support active research in this domain. Our comprehensive dataset is adapted from real-world conversations between users and experts, creating samples that are more representative of questions and images one may ask. This enables more accurate responses as demonstrated by our fine-tuning experiments. This dataset can be used to support the development of an agricultural vision language model that can provide users with instant assistance on various topics like insect/pest identification, disease categorization, and most importantly, management instructions. When properly used, these models have the potential to assist sustainability goals, prevent yield loss, and improve resource use. 
% \textless One paragraph to state that our work has very positive potential societal impact. Describe the reasons for this.\textgreater

% \newpage
% \input{sections/X1_neurips_checklist}

% \input{sections/X_supplementary}

\end{document}